\documentclass[10pt,twocolumn,letterpaper]{article}

\usepackage{cvpr}%

\usepackage{pgfplots,pgf-pie}
\usepackage{subcaption}
\usepackage{wrapfig}
\usepackage{xspace}
\newcommand{\dsetname}{FlatSounds\xspace}

\definecolor{cvprblue}{rgb}{0.21,0.49,0.74}
\usepackage[pagebackref,breaklinks,colorlinks,allcolors=cvprblue]{hyperref}
\usepackage[most]{tcolorbox}
\tcbset{
  colback=white,
  colframe=black,
  boxrule=0.3pt,
  arc=1.5pt,
  outer arc=1.5pt,
  left=4pt,
  right=4pt,
  top=4pt,
  bottom=4pt
}
\usepackage{inconsolata}
\usepackage{algorithm}
\usepackage{algpseudocode}

\def\paperID{35680}
\def\confName{CVPR}
\def\confYear{2026}
\newcommand{\mypar}[1]{\vspace{-4mm}\paragraph{{\bf #1}}}
\renewcommand{\thefootnote}{\fnsymbol{footnote}}

\title{Benchmarking Single-Factor Physical Video-to-Audio Generation}

\author{
Tingle Li$^{1,2}$\footnotemark[1] \quad
Siddharth Gururani$^2$\footnotemark[1] \quad
Kevin J. Shih$^2$\footnotemark[1] \quad
Gantavya Bhatt$^3$ \quad
Sang-gil Lee$^2$ \\
Zhifeng Kong$^2$ \quad
Arushi Goel$^2$ \quad
Gopala Anumanchipalli$^1$ \quad
Ming-Yu Liu$^2$ \\[0.3em]
$^1$UC Berkeley \quad $^2$NVIDIA \quad $^3$University of Washington\\
\small \url{https://research.nvidia.com/labs/cosmos-lab/flatsounds/}
}

\begin{document}

\twocolumn[{
\maketitle
\begin{center}
    \centering \captionsetup{type=figure} \includegraphics[width=\linewidth]{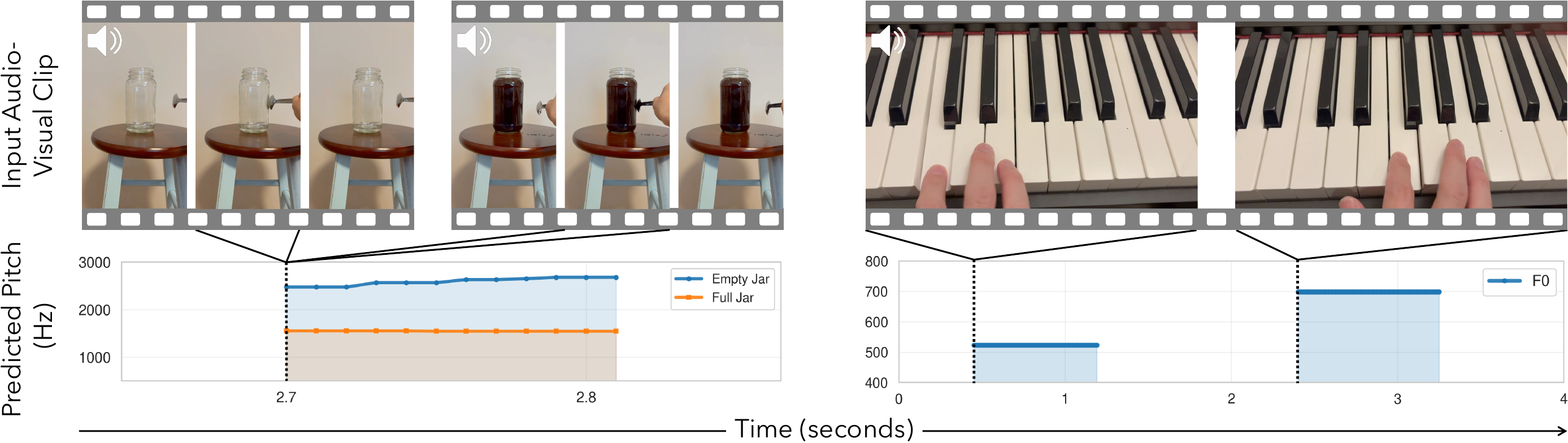} \captionof{figure}{\small {\bf FlatSounds for video-to-audio physical benchmark.} We test whether current video-to-audio models generate sound that reflects controlled changes in physical factors. For counterfactual pairs, we time-warp videos so that only a single physical factor (e.g., jar fullness) differs while impact timing remains aligned, and compare generated physical features such as pitch (left). For single-video tests, we probe within-clip consistency and directional trends, such as increasing pitch for ascending piano key presses (right).
    }
    \label{fig:teaser}
\end{center}
}]

\footnotetext[1]{Equal contribution}
\renewcommand{\thefootnote}{\arabic{footnote}}
\begin{abstract}
Generative video-to-audio (V2A) models produce highly plausible soundtracks, but it remains unclear whether they capture the underlying physical processes. Existing evaluations emphasize perceptual realism and overlook physical correctness under controlled interventions. In this paper, we introduce FlatSounds, a benchmark that audits the physical reasoning of V2A models through: 1) controlled counterfactual pairs in which a single physical factor is varied, and 2) single-video pattern tests that probe internal consistency and directional trends. These settings test whether the generated audio correctly reflects specific physical properties and timings. Our evaluation of state-of-the-art models reveals a consistent trade-off: models rely more on text captions than the visual stream to infer physics and semantics. Captions generally improve physical and semantic accuracy, but paradoxically degrade temporal alignment. Our results highlight the need to move beyond audio quality toward learning physical processes directly from pixels. Finally, we find that our physics-based metrics correlate strongly with human preference tests on our own data.
\end{abstract}
    
\vspace{-5.8mm}
\section{Introduction}
\label{sec:intro}
A core aspiration of artificial intelligence is to build general-purpose world models \cite{ha2018world, lecun2022path}, systems that can simulate, reason, and interact with the physical world. A simulation of reality, however, is incomplete if it is silent, as sound is a rich physical signal that reveals latent properties of the world, such as material, fullness, or unseen dynamics, that are often ambiguous to vision alone \cite{kunkler2000hearing, bagad2025sound, wang2025sound}. The task of video-to-audio (V2A) generation \cite{owens2016visually, zhou2018visual} serves as a critical test bed for this grand challenge. To successfully generate audio from video, a model must implicitly simulate the physical processes that produce sound. When a metal spoon strikes a glass (Fig.~\ref{fig:teaser}), the resulting acoustic texture \cite{mcdermott2011sound} is strictly governed by the object's geometry, material composition, and interaction dynamics \cite{farnell2010designing, o2002synthesizing}. A model that truly understands this event is not merely pattern matching, but also simulating an internal physics engine \cite{baillargeon2002acquisition, smith2005development}.

However, this perspective reveals a stark deficiency in the current evaluation. Recent V2A models \cite{zhang2024foleycrafter, cheng2025mmaudio, chen2025video, liu2025thinksound} produce remarkably plausible sounds, but their success is largely measured by distributional and semantic metrics (e.g., FAD and CLAP) \cite{kilgour2018fr, wu2023large} that capture surface-level {\em plausibility} rather than deep {\em physical understanding} \cite{ding2022cogview2, otani2023toward, huang2024vbench, vinay_evaluating_2022}. This method obscures a fundamental question: to what degree do these models truly capture the dynamics behind how and why a sound is emitted from physical interaction?

We argue that current benchmarks \cite{polyak2024movie, cheng2025mmaudio, wang2025kling} are insufficient because they test {\em correlation}, not {\em causal responsiveness}. Generating a plausible \textit{clink} for a glass does not imply a correct internal model of the underlying physics. We must evaluate the model's response to controlled causal interventions \cite{peters2017elements}, where individual physical factors are systematically manipulated. How does the sound change when the striker's material shifts from metal to wood, or when a container's fullness is modified \cite{cabe2000human}? A model with a robust physical world model will correctly modulate acoustic properties such as attack time or fundamental frequency \cite{zwicker2013psychoacoustics, howard2013acoustics}, while a model built on plausibility alone might fail.

To address this gap, we introduce FlatSounds, a new V2A physical evaluation framework to systematically audit the physical dimension of V2A models. Our benchmark complements conventional metrics by introducing two evaluation modes: controlled counterfactual testing and single-video pattern analysis. We curate time-warped video pairs in which a single physical factor (e.g., material, geometry, environment) is modified while other factors remain fixed, and we design single-video tests for internal consistency (e.g., repeated identical impacts) and directional trends (e.g., ascending pitch). Notably, for the counterfactual pairs, we use time-warping to precisely align impact timings, ensuring the isolated physical variable is the sole cause of any acoustic change. We then use metrics grounded in physics to measure if the change in generated audio correctly reflects the video's delta, and use timing-related metrics to measure temporal alignment.

Our evaluation of state-of-the-art V2A models \cite{zhang2024foleycrafter, cheng2025mmaudio, liu2025thinksound, shan2025hunyuanvideo} under varied conditioning settings yields a striking insight: while text captions generally improve semantic alignment and physical correctness, they paradoxically degrade temporal synchronization. Across most models, removing captions actually {\em improves} certain timing-related metrics. This exposes a critical bottleneck in the video encoder. Current models do not truly see the physics; they preferentially read it from text when available, ignoring the precise visual cues for timing and physical interaction. They effectively ``cheat'' by relying on explicit text descriptions rather than emergent visual understanding. The pronounced performance drop in physical accuracy when captions are removed confirms this dependency, highlighting a fundamental weakness in current visual representation learning for physical processes. Our key contributions are:
\begin{itemize}
    \item We introduce a benchmark that reframes V2A evaluation from plausibility to physical correctness, providing a new tool to audit whether models respond causally to controlled changes in physical factors.
    \item We curate a dataset and protocols with two complementary modes: time-warped factual–counterfactual pairs that isolate a single physical variable, and single-video consistency tests for repeated patterns and directional trends, enabling joint assessment of physical correctness and temporal alignment. 
    \item We show that current V2A models tend to rely on text for physical reasoning at the expense of visual temporal alignment, revealing a core deficiency in video encoders and reframing the central challenge as one of visual physical understanding.
\end{itemize}

\section{Related Works}
\label{sec:related}
\paragraph{Video-to-audio generation.}
The field of V2A generation has evolved from specialized models for specific interactions \cite{owens2016visually} to open-domain systems. Current methods employ diverse architectures, including auto-regressive transformer models \cite{iashin2021taming, sheffer2023hear, viertola2025temporally}, diffusion, flow-matching, and MaskGIT models for high-fidelity output \cite{luo2023diff, xu2024video, wang2024frieren, liu2024tell, pascual2024masked, su2024vision}, and frameworks for video-audio co-generation \cite{haji2025av, zhang2025uniavgenunifiedaudiovideo}, which typically involve parallel transformer blocks for the video and audio modality with some form of cross-modality fusion block. In contrast, a dominant and highly successful approach, which includes several of our evaluated baselines \cite{zhang2024foleycrafter, cheng2025mmaudio, shan2025hunyuanvideo}, involves a two-stage process: first, adapting pre-trained text-to-audio diffusion models \cite{liu2023audioldm} with visual adapters \cite{radford2021learning, iashin2024synchformer}, or employing multi-modal joint training with text-audio data \cite{cheng2025mmaudio}. This reliance on text is even more explicit in models such as ThinkSound \cite{liu2025thinksound}, which integrates a multi-modal LLM to first generate a textual chain-of-thought reasoning before synthesizing the audio. While these methods achieve state-of-the-art plausibility, their architectural reliance on text motivates our need for a benchmark that can audit whether visual-physical grounding is being ignored. Our contribution is a new benchmark to evaluate this critical dimension.

\mypar{Video-to-audio evaluation.}
Current V2A evaluation is primarily focused on plausibility, using datasets and metrics designed to test in-the-wild correlations. The most common evaluation datasets are large-scale, unconstrained collections of internet videos, such as AudioSet \cite{gemmeke2017audio}, VGGSound \cite{chen2020vggsound}. These are essential for training, but as uncontrolled collections, they do not provide the paired ground truth needed for causal intervention analysis. Furthermore, creating the necessary controlled interventions post-hoc via video manipulation remains an unsolved challenge \cite{xing2024survey}. The primary evaluation metrics fall into two main categories. First are distributional and semantic metrics, such as FAD \cite{kilgour2018fr}, CLAP \cite{wu2023large}, and ImageBind \cite{girdhar2023imagebind}. Second are temporal synchronization metrics, which typically use specialized models like Synchformer \cite{iashin2024synchformer} to estimate the weighted temporal offset. Comprehensive benchmark suites, such as AV-Benchmark \cite{cheng2025mmaudio}, Movie Gen Audio Bench \cite{polyak2024movie}, and Kling-Audio-Eval \cite{wang2025kling}, package these semantic and temporal metrics into robust toolkits for evaluating broad performance. While this landscape is mature, it is designed to measure plausibility and correlation. Our work fills a critical gap by providing a new framework to answer a causal question using controlled interventions.

\mypar{Physical and counterfactual evaluation.}
Recent work in evaluating generative models has diverged into two key trends. The first focuses on physical reasoning benchmarks across visual domains. Recognizing the limits of plausibility, frameworks have emerged to audit physical understanding in video generation models. These include VLM-based assessments \cite{bansal2024videophy, meng2024towards}, object tracking metrics \cite{wang2024you}, human-aligned metrics \cite{huang2024vbench}, object-level specificity \cite{motamed2025generative, agarwal2025cosmos}, and real physics experiments \cite{zhang2025morpheus}. In parallel, for vision-language models, benchmarks like PhysBench \cite{chow2025physbench} and PAI Bench \cite{zhou2025pai} audit an understanding of concepts like mass and density. We extend this line of inquiry from visual properties to the audio domain. The second trend is causal and counterfactual reasoning \cite{peters2017elements}. This framework is increasingly used to diagnose and correct ``shortcut learning'' \cite{geirhos2020shortcut}. Benchmarks in visual question answering \cite{johnson2017clevr, abbasnejad2020counterfactual} use counterfactual interventions to force models to learn true causal structures. Our work proposes to synthesize these two threads: we audit acoustic-physical properties using a counterfactual intervention framework.

\mypar{Acoustics.}
The link between the physical properties of an object and its sound is well-established \cite{gaver1993world, bregman1994auditory}. Acoustics is a broad field of physics that provides formal models linking an object's physical attributes, such as geometry, material stiffness, and boundary conditions, to its resulting sound texture \cite{mcdermott2011sound}, including its modal resonances (which determine pitch) and high-frequency damping (which determines timbre) \cite{farnell2010designing, o2002synthesizing}. On the perceptual side, psychoacoustics studies the human perception of these physical-acoustic links \cite{blauert1997spatial, moore2012introduction}. A key concept in this field is the Just Noticeable Difference (JND) \cite{zwicker2013psychoacoustics, del2022study}, which defines the perceptual threshold for changes in features such as fundamental frequency (pitch) or decay rate (duration). Prior work in this area has specifically characterized the perceptual salience of features such as attack time, which strongly correlates with human perception of material hardness (e.g., metal and wood) \cite{freed1990auditory}. Our benchmark leverages this body of work by selecting a suite of objective, perceptually-relevant acoustic measures for evaluating a model's physical understanding, moving beyond correlation-based scores.

\section{Benchmark Metrics}
\label{sec:benchmark}
We evaluate two complementary aspects of V2A models: 1) temporal alignment, and 2) whether the generated sound responds correctly to single-factor physical changes in the conditioning video.

\subsection{Temporal Alignment}
\label{sec:alignment}
We measure audio-video alignment on impact-style events. Our benchmark videos heavily feature actions such as
tapping, scratching, plucking, clapping, and smacking. These actions typically produce sounds with short attacks and distinct peaks in their amplitude envelope. We manually annotate the onset times of such events in each video (see Sec. \ref{sec:hit_anno} for details).

Given the annotated event times, we evaluate the generated audio by detecting onset candidates using an onset-strength detector with an envelope-based fallback, and checking whether at least one detected event lies within an adaptive temporal window around each annotation. This yields an event recall score that we term Hit Coverage (\%). We focus on recall rather than precision, since additional unannotated sounds (e.g., ambient noise) may be physically plausible and should not be penalized. We also measure the Timing Error as the average deviation (in ms) of detected hits from the ground truth annotation. Finally, since we generate multiple audio samples per video, we define Perfect Align as the percentage of these generations that achieve 100\% Hit Coverage.

\subsection{Physical Correctness}
\label{sec:correctness_attributes}
Instead of requiring absolute accuracy in pitch ranges or attack times, we focus on directional consistency under controlled physical interventions. When a single relevant factor in the video is changed, we ask whether the corresponding change in the generated audio has the expected direction. For instance, if a clapping video is moved from a heavily furnished living room to a wide stairwell, we expect an increase in reverberation. Our benchmark realizes such single-factor changes across several sound attributes (see visualizations in the supplement) and measures whether the predicted attribute changes in the correct direction.

\mypar{Temporal envelope features.}
We capture impact dynamics and material-dependent damping through three envelope-level quantities:
\begin{itemize}
\item \textbf{Attack Time}: The time it takes a sound to reach its peak amplitude after onset. Replacing a soft material with a harder one (e.g., foam with metal) decreases attack time.
\item \textbf{Decay Rate}: How quickly a sound decays after the attack. A sheet of metal in firm contact with a table is more strongly damped and should exhibit a higher decay rate than the same sheet freely hanging in the air.
\item \textbf{Temporal Modulation}: The strength of rhythmic or amplitude fluctuations over time. Shaking a jar of coins should yield stronger temporal modulation than shaking a jar of sand.
\end{itemize}

\begin{figure*}
    \centering
    \includegraphics[width=\linewidth]{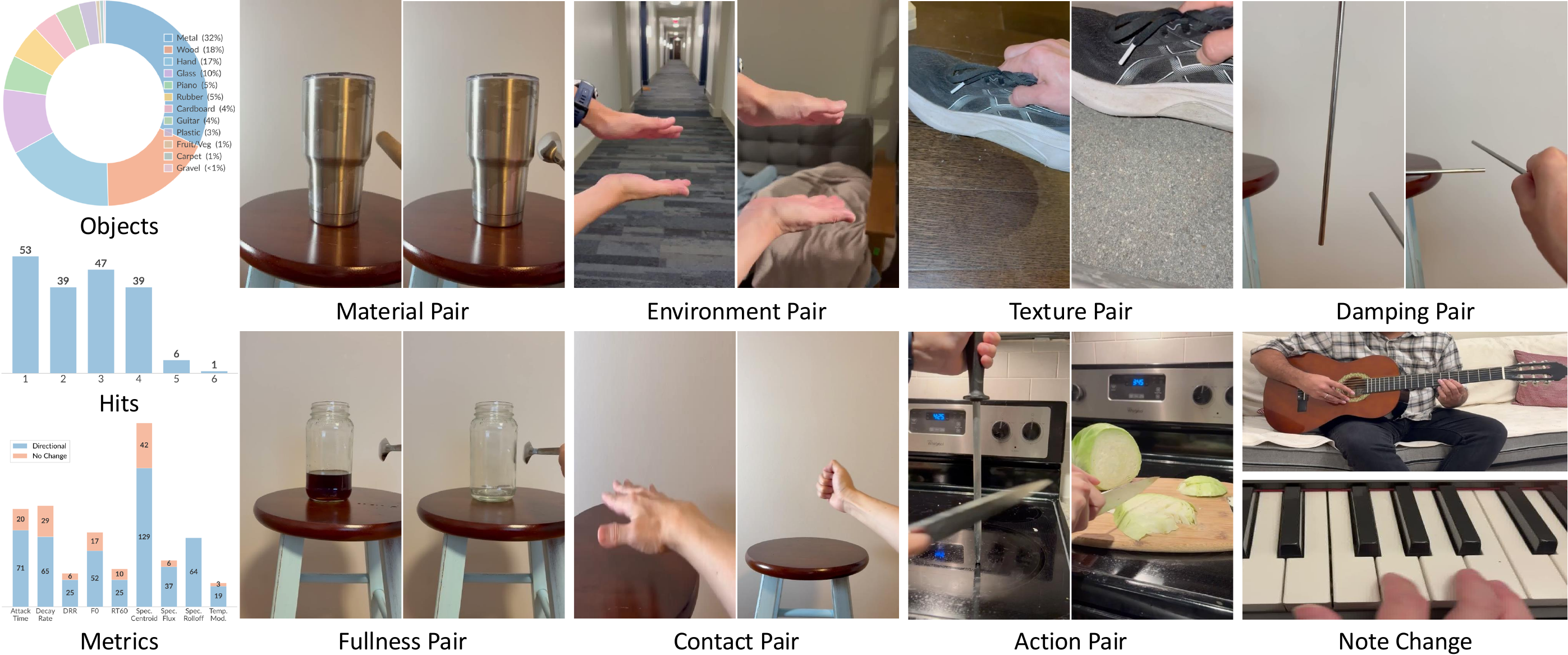}
    \caption{{\bf \dsetname dataset.} How should sound change when we manipulate specific visual properties of a scene? \dsetname\ contains indoor recordings of everyday household objects producing sound under controlled variations. Many clips are arranged into time-aligned counterfactual pairs in which we alter a single factor (material, environment, texture, etc.) while keeping the rest of the scene fixed, allowing us to test whether generated audio changes in the expected way. We show the distribution of objects, hits, and metrics (left), as well as example frames of several counterfactual pairs presented in the dataset (right).}
    \label{fig:data}
\end{figure*}

\mypar{Room acoustics.}
We measure changes in environment using two standard reverberation measures:
\begin{itemize}
\item \textbf{Room Reverberation Time (RT60)}: The time it takes the sound level to decay by 60dB. The RT60 of a large hall is typically on the order of several seconds, while a small absorptive room has an RT60 below one second.
\item \textbf{Direct-to-Reverberant Sound Ratio (DRR)}: The ratio between the energy of the direct sound and the energy of the reverberant field. It quantifies the clarity of an environment: a large hall has a lower DRR than a small room since the reverberant energy is relatively stronger.
\end{itemize}

\mypar{Spectral and tonal structure.}
We probe material and excitation changes via spectral and pitch descriptors:
\begin{itemize}
\item \textbf{Fundamental Frequency (F0)}: The lowest-frequency component of a periodic sound. It is the primary physical correlate of perceived pitch~\cite{cheveigne2010pitch}. F0 should ascend or descend as we play ascending or descending notes on any instrument.
\item \textbf{Spectral Centroid}: The center of gravity of the magnitude spectrum in each STFT frame.
The spectral centroid is perceived as the sound's ``brightness'' or ``sharpness''~\cite{Bismarck1974SharpnessAA, iverson1993isolating}. Harder materials should have an increased spectral centroid~\cite{freed1990auditory,giordano2006material}. For example, an impact on glass should produce a higher centroid than a similar impact on wood or on a sofa.
\item \textbf{Spectral Flux}: The average frame-to-frame change in spectral magnitude~\cite{lerch_introduction_2023}. Spectral flux is related to the perceived ``roughness'' or ``busyness'' of a sound~\cite{zwicker2013psychoacoustics}. For instance, smoothly cutting paper with scissors should exhibit lower spectral flux than tearing paper.
\item \textbf{Spectral Rolloff}: The frequency below which a fixed proportion (typically 85\%) of the total spectral energy is contained~\cite{lerch_introduction_2023}. Stepping on dry leaves should yield a higher rolloff frequency than stepping on wet leaves, due to the wider spread of energy at higher frequencies.
\end{itemize}

\section{Benchmark Construction}
\label{sec:collection}
We describe the construction of \dsetname, the video dataset that supports our alignment and physical correctness benchmarks. The dataset is deliberately compact but tightly controlled: each clip is recorded with a specific physical factor and metric in mind, enabling targeted counterfactual and single-video tests rather than uncontrolled coverage.

\subsection{Data Collection}
\dsetname\ comprises 185 unique indoor clips (Fig.~\ref{fig:data}), all of which are used in both our single-video alignment analysis and our physical correctness evaluations. Each clip is paired with a human-written caption and timestamps of sound-producing events.

We design the following criteria for collecting video clips: 1) each clip should be between 5 and 10 seconds; 2) the sound-producing event should be roughly centered within the frame; 3) the interaction produces clear, high-energy onsets with a distinct peak in the audio energy envelope; 4) sound events should occur roughly 1-5 times per video, with at least half a second in between each event to facilitate event detection. We focus on impact-style and other clear-onset events because their discrete timing and relatively clean causal structure make single-factor attribution and temporal-alignment auditing more reliable.

All videos are recorded indoors using smartphone cameras and onboard microphones. We focus on everyday physical interactions between household objects (jars, utensils), furniture and structural elements (doors, floors), human hands, and instruments. To probe room acoustics, we record matched interactions in acoustically distinct environments, such as near cushioned sofas \vs hard walls, and in narrow corridors or stairwells \vs more absorptive living spaces.

The clips are organized around the metrics introduced in Sec.~\ref{sec:benchmark}. For temporal metrics, we record impact and short-friction events that differ in material, contact geometry, or damping (e.g., the same metal object struck on a table \vs hanging freely). For spectral metrics, we design clips that systematically vary pitch, material hardness, or texture (e.g., gravel \vs wood impacts). For environmental metrics, we record the same gesture across rooms that differ in size and absorption. Many of these clips are further grouped into factual-counterfactual pairs that isolate a single manipulated factor (material, fullness, environment, or action), while others are used as single-video pattern tests (e.g., monotonic pitch sequences or repeated identical hits). We refer to this benchmark as \dsetname\ to emphasize its focus on everyday physics within an apartment-like setting.

\subsection{Annotation}
\label{sec:hit_anno}
Each video is annotated with a text caption and exact timings for sound events of interest. In order to introduce a degree of automation into this process, we narrow our set of sound-generating events to ones that produce a distinct peak when plotting the sound energy envelope. This limits us to common impact-generating sounds such as clapping, tapping, playing musical notes, and fast friction events.

\mypar{Sound event timing annotation.}
To obtain precise event timings, we follow a semi-automatic procedure. For each audio track, we compute a short-time Fourier transform and derive a frame-wise energy envelope by taking the RMS of the magnitude spectrum across frequencies. We then apply \texttt{scipy.find\_peaks}~\cite{2020SciPy-NMeth} with conservative detection thresholds to get candidate timestamps of onset peaks. In practice, background noise and long-tailed resonances, especially from ringing metal, often introduce multiple false-positive peaks (see the supplementary material for details). We therefore perform a manual verification step: inspect each candidate peak, remove spurious detections, add any missed onsets, and discard clips for which a clean set of event peaks cannot be reliably extracted. The resulting verified timestamps serve as ground truth for our alignment metrics and as anchor points for constructing time-aligned counterfactual pairs.

\mypar{Creating counterfactual pairs.} 
We construct counterfactual video pairs to evaluate the model's causal responsiveness. Our goal is to temporally align the sound events to the best of our ability, such that the video pairs only differ in the object, action, or background setting that we alter.

We identify pairs from \dsetname using ground-truth audio annotations and expert knowledge of acoustics, annotating each pair with at least one metric change from Sec.~\ref{sec:correctness_attributes}. At test time, we check that the generated models comply with the direction of the annotated metric change (e.g., we expect an increase in Temporal Modulation from video A to B). The counterfactual video must contain at least as many annotated sound events (hits) as the factual video. We use only the first $N$ hits for alignment if there are more. We record multiple versions of each interaction with varying hit counts to enable flexible pairing.

For temporal alignment, we warp the counterfactual video to match the factual video's event timings. We set time-annotated sound events as anchor points, then stretch or compress the time between anchor points such that the annotated peaks occur at the target timestamps. Frames between anchor points are resampled to achieve the target frame count and segment duration. The time-warping procedure may result in unnatural motion if timestamps differ substantially, but we find results generally good without introducing visual artifacts. For single sound events, warping is unnecessary as we can achieve alignment through frame-trimming alone.

\section{Experiments}
\label{sec:exp}
Here, we evaluate recent V2A models using our benchmark, as well as existing metrics such as FAD, KL, Inception Score, ImageBind, CLAP, DeSync. We aim to demonstrate that our \dsetname benchmark and physics-informed metrics evaluate V2A model quality in an interpretable and effective manner, and would serve as a good complement to existing evaluation frameworks.

\begin{table*}[t]
\centering
\scriptsize
\setlength{\tabcolsep}{4pt}
\caption{Per-metric average confidence on \dsetname-Physics. ``Temporal Mod.'' denotes Temporal Modulation, and ``Avg.'' reports the average confidence over all metrics. All metrics are higher-is-better ($\uparrow$).}
\label{tab:pav_physics}
\begin{tabular}{lccccccccccc}
\toprule
\textbf{Method} &
\textbf{Attack Time} &
\textbf{Decay Rate} &
\textbf{F0} &
\textbf{Spectral Centroid} &
\textbf{Spectral Flux} &
\textbf{Spectral Rolloff} &
\textbf{Temporal Mod.} &
\textbf{RT60} &
\textbf{DRR} &
\textbf{Avg.} \\
\midrule
MMAudio-Phys (w/ Caption)  & 0.290 & 0.310 & \textbf{0.334} & \textbf{0.368} & 0.321 & \textbf{0.395} & 0.237 & 0.310 & 0.189 & \textbf{0.306} \\
Hunyuan-V2A (w/ Caption)   & 0.267 & \textbf{0.320} & 0.326 & 0.332 & \textbf{0.403} & 0.305 & 0.283 & 0.247 & 0.262 & 0.305 \\
Hunyuan-V2A (w/o Caption)  & 0.232 & 0.306 & 0.258 & 0.294 & 0.383 & 0.300 & 0.320 & 0.271 & \textbf{0.300} & 0.296 \\
MMAudio-Phys (w/o Caption) & \textbf{0.293} & 0.279 & 0.310 & 0.318 & 0.309 & 0.347 & 0.263 & \textbf{0.316} & 0.164 & 0.289 \\
ThinkSound (w/ Caption)    & 0.243 & 0.188 & 0.198 & 0.267 & 0.174 & 0.294 & 0.299 & 0.232 & 0.157 & 0.228 \\
MMAudio (w/ Caption)       & 0.209 & 0.175 & 0.228 & 0.259 & 0.253 & 0.223 & 0.266 & 0.232 & 0.189 & 0.226 \\
MMAudio (w/o Caption)      & 0.200 & 0.171 & 0.230 & 0.265 & 0.237 & 0.243 & 0.196 & 0.217 & 0.230 & 0.221 \\
ThinkSound (w/o Caption)   & 0.213 & 0.194 & 0.249 & 0.266 & 0.246 & 0.248 & 0.225 & 0.177 & 0.152 & 0.219 \\
FoleyCrafter (w/o Caption) & 0.180 & 0.179 & 0.182 & 0.241 & 0.294 & 0.215 & 0.299 & 0.176 & 0.188 & 0.217 \\
FoleyCrafter (w/ Caption)  & 0.158 & 0.157 & 0.177 & 0.234 & 0.267 & 0.230 & \textbf{0.333} & 0.143 & 0.147 & 0.205 \\
\bottomrule
\vspace{-2em}
\end{tabular}
\end{table*}

\begin{table}[t]
\centering
\scriptsize
\setlength{\tabcolsep}{1.7pt}
\caption{Overall performance on \dsetname-Physics. All metrics are higher-is-better ($\uparrow$).}
\label{tab:overall_summary}
\begin{tabular}{lcccc}
\toprule
\textbf{Method} &
\textbf{Confidence} &
\textbf{Hit Coverage} &
\textbf{Perfect Align} &
\textbf{CLAP} \\
\midrule
MMAudio-Phys (w/ Caption)  & \textbf{0.306} & 82.65 & 59.82 & 0.630 \\
Hunyuan-V2A (w/ Caption)   & 0.305 & 90.21 & 69.31 & 0.633 \\
Hunyuan-V2A (w/o Caption)  & 0.296 & \textbf{91.50} & \textbf{70.50} & 0.593 \\
MMAudio-Phys (w/o Caption) & 0.289 & 83.69 & 61.00 & 0.602 \\
ThinkSound (w/ Caption)    & 0.228 & 74.81 & 51.52 & 0.573 \\
MMAudio (w/ Caption)       & 0.226 & 75.02 & 52.03 & \textbf{0.642} \\
MMAudio (w/o Caption)      & 0.221 & 75.81 & 51.50 & 0.616 \\
ThinkSound (w/o Caption)   & 0.219 & 75.05 & 51.10 & 0.548 \\
FoleyCrafter (w/o Caption) & 0.217 & 83.39 & 60.70 & 0.548 \\
FoleyCrafter (w/ Caption)  & 0.205 & 66.52 & 44.70 & 0.573 \\
\bottomrule
\end{tabular}
\end{table}

\subsection{Experimental Setup}
\paragraph{Evaluated models.}
We evaluate a representative set of state-of-the-art V2A models: FoleyCrafter \cite{zhang2024foleycrafter}, Hunyuan-V2A \cite{shan2025hunyuanvideo}, MMAudio \cite{cheng2025mmaudio}, ThinkSound \cite{liu2025thinksound}. To test the hypothesis that models rely on explicit textual cues for physics, we also include MMAudio-Phys, our own fine-tuned variant of MMAudio using physics-aware captions gathered using custom prompts with Omni-captioner \cite{ma2025omni}. We also evaluate the contribution of the caption during inference by evaluating all models with and without captions.

\mypar{Datasets and evaluation protocols.}
We evaluate the available models on both VGGSound \cite{chen2020vggsound} and \dsetname. On VGGSound, we employ all standard evaluation protocols: Fréchet Audio Distance (FAD, computed with PANN, PaSST, and VGG embeddings) \cite{kilgour2018fr}, KL, Inception Score (IS) \cite{salimans2016improved}, ImageBind Score (IB) \cite{girdhar2023imagebind}, CLAP score \cite{wu2023large}, and DeSync \cite{iashin2024synchformer}. On \dsetname, we run all aforementioned standard evaluations in addition to our proposed metrics for alignment and physical correctness, as described in Sec.~\ref{sec:benchmark}.

\begin{table*}[t]
\centering
\scriptsize
\caption{Comparison of V2A models on VGGSound and standard benchmark metrics. All models are evaluated under two conditions: with (w/) and without (w/o) text captions. $\downarrow$ indicates lower is better. $\uparrow$ indicates higher is better. Best results are in bold.}
\label{tab:vggsound_comp}
\begin{tabular}{lcccccccc}
\toprule
\textbf{Method} &
\textbf{FAD-PASST $\downarrow$} &
\textbf{FAD-PANN $\downarrow$} &
\textbf{FAD-VGG $\downarrow$} &
\textbf{KL-PANNS $\downarrow$} &
\textbf{KL-PASST $\downarrow$} &
\textbf{IS $\uparrow$} &
\textbf{IB $\uparrow$} &
\textbf{DeSync $\downarrow$} \\
\midrule
MMAudio-Phys (w/ Caption)   & 54.73 & \textbf{3.97} & \textbf{0.61} & \textbf{1.37} & \textbf{1.16} & \textbf{18.49} & \textbf{34.89} & 0.405 \\
MMAudio-Phys (w/o Caption)  & 66.31 & 4.38 & 0.64 & 1.75 & 1.49 & 13.43 & 33.11 & 0.399 \\
MMAudio (w/ Caption)        & 65.86 & 4.89 & 1.08 & 1.68 & 1.40 & 17.59 & 33.03 & 0.445 \\
MMAudio (w/o Caption)       & 63.84 & 4.50 & 0.80 & 2.12 & 1.82 & 14.45 & 32.46 & 0.436 \\
FoleyCrafter (w/ Caption)   & 182.46 & 18.45 & 3.06 & 2.25 & 2.24 & 15.31 & 25.87 & 1.195 \\
FoleyCrafter (w/o Caption)  & 191.30 & 19.53 & 3.82 & 2.50 & 2.42 & 10.89 & 28.27 & 1.172 \\
Hunyuan-V2A (w/ Caption)        & 78.38 & 10.02 & 2.11 & 2.04 & 1.79 & 15.29 & 31.66 & 0.340 \\
Hunyuan-V2A (w/o Caption)       & 114.51 & 15.06 & 2.12 & 2.52 & 2.28 & 7.59 & 31.19 & \textbf{0.326} \\
ThinkSound (w/ Caption)     & {\bf 52.44} & 4.82 & 0.71 & 1.44 & 1.26 & 17.98 & 29.24 & 0.455 \\
ThinkSound (w/o Caption)    & 62.32 & 5.02 & 0.75 & 1.64 & 2.18 & 10.30 & 26.74 & 0.433 \\
\bottomrule
\vspace{-3em}
\end{tabular}
\end{table*}

\begin{table}[t]
\centering
\scriptsize
\caption{Temporal comparison on \dsetname-Single, where Hit Coverage (\%) and Timing Error (ms) are reported.}
\label{tab:pav_temporal}
\begin{tabular}{lcc}
\toprule
\textbf{Method} & \textbf{Hit Coverage} $\uparrow$ & \textbf{Timing Error} $\downarrow$\\
\midrule
Ground Truth                 & 97.12 $\pm$ 1.72 & 17.25 $\pm$ 2.64 \\
\midrule
Hunyuan-V2A (w/o Caption)   & \textbf{68.55 $\pm$ 3.52} & \textbf{44.34 $\pm$ 1.04} \\
Hunyuan-V2A (w/ Caption)    & 65.21 $\pm$ 3.81 & 44.76 $\pm$ 1.01 \\
MMAudio-Phys (w/o Caption)  & 56.46 $\pm$ 2.77 & 46.63 $\pm$ 1.05 \\
MMAudio-Phys (w/ Caption)   & 50.69 $\pm$ 4.23 & 51.34 $\pm$ 1.09 \\
FoleyCrafter (w/o Caption)  & 49.74 $\pm$ 4.25 & 49.32 $\pm$ 1.09 \\
FoleyCrafter (w/ Caption)   & 48.85 $\pm$ 3.07 & 51.48 $\pm$ 1.12 \\
ThinkSound (w/o Caption)    & 36.34 $\pm$ 3.58 & 53.15 $\pm$ 1.19 \\
ThinkSound (w/ Caption)     & 33.74 $\pm$ 3.61 & 53.66 $\pm$ 1.21 \\
MMAudio (w/o Caption)       & 31.95 $\pm$ 3.88 & 56.20 $\pm$ 1.17 \\
MMAudio (w/ Caption)        & 31.12 $\pm$ 3.85 & 57.67 $\pm$ 1.20 \\
\bottomrule
\end{tabular}
\end{table}

\begin{table*}[t]
\centering
\scriptsize
\setlength{\tabcolsep}{2.3pt}
\caption{Comparison across V2A models on \dsetname-Single, where all metrics are presented with 95\% confidence intervals.}
\label{tab:pav_semantic}
\begin{tabular}{lcccccccc}
\toprule
\textbf{Method} &
\textbf{FAD-PASST $\downarrow$} &
\textbf{FAD-PANN $\downarrow$} &
\textbf{FAD-VGG $\downarrow$} &
\textbf{KL-PANNS $\downarrow$} &
\textbf{KL-PASST $\downarrow$} &
\textbf{IS $\uparrow$} &
\textbf{IB $\uparrow$} &
\textbf{DeSync $\downarrow$} \\
\midrule
MMAudio-Phys (w/ Caption)  & 861.47 $\pm$ 10.80 & \textbf{64.17 $\pm$ 1.51}  & 7.81 $\pm$ 0.23  & \textbf{2.13 $\pm$ 0.07} & 2.65 $\pm$ 0.07 & 4.60 $\pm$ 0.13 & 0.297 $\pm$ 0.005 & 0.723 $\pm$ 0.019 \\
MMAudio-Phys (w/o Caption) & 919.67 $\pm$ 10.69 & 73.21 $\pm$ 1.56  & 7.91 $\pm$ 0.21  & 2.48 $\pm$ 0.08 & 2.89 $\pm$ 0.07 & 4.67 $\pm$ 0.15 & 0.287 $\pm$ 0.004 & 0.717 $\pm$ 0.019 \\
MMAudio (w/ Caption)       & 865.95 $\pm$ 10.02 & 66.74 $\pm$ 1.59  & 9.29 $\pm$ 0.25  & 2.30 $\pm$ 0.08 & 2.82 $\pm$ 0.07 & 4.89 $\pm$ 0.18 & 0.313 $\pm$ 0.004 & 0.715 $\pm$ 0.018 \\
MMAudio (w/o Caption)      & 894.99 $\pm$ 10.32 & 73.65 $\pm$ 1.68  & 9.32 $\pm$ 0.24  & 2.57 $\pm$ 0.09 & 2.92 $\pm$ 0.07 & 4.89 $\pm$ 0.14 & 0.301 $\pm$ 0.005 & 0.711 $\pm$ 0.017 \\
FoleyCrafter (w/ Caption)  & 870.77 $\pm$ 9.32  & 103.48 $\pm$ 2.31 & 12.38 $\pm$ 0.34 & 3.18 $\pm$ 0.09 & 2.50 $\pm$ 0.07 & 4.62 $\pm$ 0.15 & 0.277 $\pm$ 0.004 & 1.169 $\pm$ 0.021 \\
FoleyCrafter (w/o Caption) & 978.88 $\pm$ 8.33  & 102.11 $\pm$ 1.75 & 13.94 $\pm$ 0.56 & 3.65 $\pm$ 0.09 & 3.17 $\pm$ 0.08 & \textbf{5.16 $\pm$ 0.21} & 0.255 $\pm$ 0.003 & 1.152 $\pm$ 0.021 \\
Hunyuan-V2A (w/ Caption)   & \textbf{833.82 $\pm$ 11.33} & 76.33 $\pm$ 1.67  & 8.51 $\pm$ 0.23  & 2.44 $\pm$ 0.09 & \textbf{2.40 $\pm$ 0.07} & 4.91 $\pm$ 0.14 & \textbf{0.315 $\pm$ 0.004} & 0.683 $\pm$ 0.018 \\
Hunyuan-V2A (w/o Caption)  & 893.03 $\pm$ 10.07 & 80.55 $\pm$ 1.67  & 8.96 $\pm$ 0.24  & 2.87 $\pm$ 0.09 & 2.84 $\pm$ 0.07 & 4.94 $\pm$ 0.15 & 0.299 $\pm$ 0.004 & \textbf{0.679 $\pm$ 0.019} \\
ThinkSound (w/ Caption)    & 972.98 $\pm$ 13.71 & 77.58 $\pm$ 1.45  & \textbf{6.89 $\pm$ 0.22}  & 2.70 $\pm$ 0.08 & 2.86 $\pm$ 0.07 & 4.93 $\pm$ 0.11 & 0.235 $\pm$ 0.004 & 0.754 $\pm$ 0.019 \\
ThinkSound (w/o Caption)   & 1044.44 $\pm$ 11.34 & 93.80 $\pm$ 1.57 & 8.22 $\pm$ 0.27  & 3.30 $\pm$ 0.08 & 3.22 $\pm$ 0.07 & 4.98 $\pm$ 0.16 & 0.210 $\pm$ 0.004 & 0.760 $\pm$ 0.019 \\
\bottomrule
\vspace{-3em}
\end{tabular}
\end{table*}

\subsection{\dsetname Metric Implementation}
\paragraph{Temporal alignment metrics.} 
Our temporal alignment metrics (Sec.~\ref{sec:alignment}) evaluate whether the generated audio from a given video produces energy peaks at the annotated timestamps extracted from the ground-truth audio. These metrics are computed on all 185 captioned and timing-annotated clips (\dsetname-Single).

\mypar{Physical correctness metrics.}
Our physical correctness metrics operate on a set constructed from 185 unique indoor clips referred to as \dsetname-Physics. Using time-warped pair construction, we get 178 paired video tests containing at least one annotated metric with an expected direction of change ($|\Delta|$), and 90 single-video tests for physical metrics that do not require counterfactual pairs (internal consistency and directional trends within a video). This gives us a total of 268 test cases for \dsetname-Physics.

It would be meaningless to judge directional changes in metrics if the generated audio is too far from the expected content to even compare. We alleviate this with a ``soft gate'' to ensure audio is both temporally aligned and semantically plausible \emph{before} judging its physical correctness. The final Confidence score for any metric is computed from per-seed weighted votes. For each seed, we define a quality weight that equally balances temporal alignment and semantic plausibility; for pair comparisons, the temporal and semantic terms each take the minimum of the factual and counterfactual scores for that seed. Audio that is poorly synchronized or semantically incorrect is not discarded but has its influence on the final score proportionally reduced. Finally, we report Confidence for each metric as the weighted proportion of seeds that satisfy the expected physical trend. We provide a detailed formulation in the supplement.
 
\mypar{Statistical testing.}
Our benchmark's Confidence score is derived from statistical voting over 10 seeds.
\begin{itemize}
    \item {\bf Increase/Decrease}: For pair comparison, a seed only votes if its delta $|\Delta|$ exceeds a robust effect-size threshold, $\tau = \max(2\% \text{ of mean}, 25\% \text{ of robust\_std})$. Seeds with sub-threshold changes ($|\Delta| \leq \tau$) or NaN values (e.g., no hit detected) are counted as failures.

    \item {\bf Ascending/Descending}: For single-video monotonicity tests (e.g., piano scale), we extract per-hit F0 sequences in log$_2$ space. For $n=2$ hits, we check the sign of the difference. For $n \geq 3$, we use Spearman's rank correlation $\rho$ with adaptive thresholds ($|\rho| \geq 0.40$ for $n \leq 4$, $0.30$ for $5 \leq n \leq 7$, and $0.25$ for longer sequences). Seeds with insufficient valid hits ($n < 2$) are counted as failures.
    \item {\bf No Change}: For pair comparison, we use a strict equivalence test: the 95\% CI of the mean delta must fall entirely within a wider band $[-\tau_{eq}, +\tau_{eq}]$. For single-video tests, we compute a robust coefficient of variation against a per-metric threshold based on JND \cite{zwicker2013psychoacoustics}.
\end{itemize}

\subsection{Physical Benchmark Performance}
We first audit models using \dsetname-Physics, which isolates causal reasoning by ensuring identical event timings.

\mypar{Physical correctness evaluation.}
The primary finding from \dsetname-Physics is that all current models struggle with physical reasoning. As summarized in Table \ref{tab:overall_summary}, the physical Confidence score remains low across models, indicating that current models struggle to learn acoustic physics from pixels. Per-metric analysis (Table \ref{tab:pav_physics}) shows MMAudio-Phys (w/ Caption) highest on average, indicating that physics-aware captions provide a noticeable boost. Most models show higher Confidence with captions, while FoleyCrafter is the only exception, with captions slightly reducing Confidence.

\mypar{Difficulty of physical properties.}
Our per-metric analysis reveals a consistent difficulty ordering. On average across models, spectral properties such as Spectral Flux, Spectral Centroid, and Spectral Rolloff are the easiest metrics, while DRR is the most challenging, with Decay Rate and Attack Time also remaining difficult. This suggests that current models capture frequency-domain characteristics more readily than fine-grained temporal dynamics and some acoustic environment cues.

\subsection{Semantic Plausibility Evaluation}
\label{subsec:standard}
Next, we evaluate all models on semantic plausibility metrics from the standard benchmark. These results, gathered from both VGGSound and the \dsetname-Single set of videos, establish a core trade-off inherent in current architectures.

As shown in Table \ref{tab:vggsound_comp}, adding captions typically improves semantic plausibility. MMAudio-Phys (w/ Caption) attains the best results on most semantic metrics, with ThinkSound (w/ Caption) achieving the best FAD-PASST. This trend holds among most models: providing text often lowers FAD and KL scores, while raising IS and IB scores. This is mirrored in our \dsetname-Single (Table \ref{tab:pav_semantic}), where captions generally reduce FAD and KL, and also improve IB, while IS does not improve consistently. This result is intuitive: explicit text guidance aids the model in generating semantically correct and high-fidelity audio.

\subsection{Temporal Alignment Evaluation}
We evaluate all models for temporal alignment with respect to videos, using our temporal alignment metrics on \dsetname videos, and DeSync on both \dsetname and VGGSound. In all cases, we observe that captions tend to degrade temporal synchronization. On VGGSound (Table \ref{tab:vggsound_comp}), Hunyuan-V2A (w/o Caption) yields the best DeSync overall, and for every model DeSync is lower (better) without captions. This pattern is corroborated on our \dsetname-Single (Table \ref{tab:pav_temporal}), where Hit Coverage is consistently higher without captions. Timing Error is also consistently lower without captions for all models, further confirming that text competes with precise visual timing.

\subsection{Human Evaluation}
Finally, we conduct a human study on a subset of our benchmark data to validate our metrics. The study uses
\setlength{\intextsep}{4pt}%
\setlength{\columnsep}{9pt}%
\begin{wraptable}{r}{0.33\columnwidth}

\scriptsize
\caption{Spearman rank-correlation of V2A metrics {\em wrt.} ELO ranking.}
\label{tab:spearman_elo}
\begin{tabular}{l c}
\toprule
\textbf{Metric} & \textbf{Value} \\
\midrule
FAD-PASST        & 0.7 \\
FAD-PANN         & 0.5 \\
FAD-VGG          & 0.6 \\
KL-PANNS         & 0.5 \\
KL-PASST         & 0.3 \\
IS               & 0.3 \\
IB               & 0.5 \\
DeSync           & 0.7 \\
CLAP             & 0.2 \\
\hline
Confidence       & \textbf{0.9} \\
Hit Coverage     & \textbf{0.9} \\
Perfect Align    & \textbf{0.9} \\
\bottomrule
\end{tabular}
\end{wraptable}\ignorespacesafterend
40 videos, sampled to broadly cover the main recording settings in our benchmark. We set up the evaluation as a pairwise preference test: each trial presents a randomly selected source video together with two generated audio tracks from two randomly selected models. The raters are asked to select the preferred video-audio pair, taking into account synchronization, presence of audio hallucinations, and overall physical plausibility. Due to the significant cost of pairwise comparisons, we limit our human study to captioned versions of MMAudio-Phys, MMAudio, FoleyCrafter, Hunyuan-V2A, and ThinkSound. The final results of the human study are summarized in Table \ref{tbl:human_elo}, where we see broad agreement between the human-study ranking and the ranking induced by our benchmark when restricted to the captioned models in Table \ref{tab:overall_summary}. Using the ELO ranking as an oracle, we calculated the Spearman rank correlation between each metric as computed on \dsetname videos, reporting the absolute value (higher is better) in Table \ref{tab:spearman_elo}. We find that our \dsetname metrics (Confidence, Hit Coverage, Perfect Align) all correlated strongly, with the best-performing standard metrics being FAD-PASST and DeSync. This result suggests that our metrics, as measured on our \dsetname data, may be more effective than most other standard metrics, while being interpretable and quick to compute. Notably, our alignment metrics correlated more strongly than DeSync. This suggests that \dsetname may be more effective for identifying alignment issues, though our metrics may not easily extend to arbitrary videos without the ability to cleanly identify sound events.
\begin{table}[t]
\centering
\scriptsize
\caption{ELO rating summary of our pairwise human evaluation study. Results are shown in descending order of rating, which largely agrees with the ordering induced by \dsetname-Physics in Table \ref{tab:overall_summary}. Full pairwise winrates are provided in the supplementary material.}
\label{tbl:human_elo}
\begin{tabular}{lc}
\toprule
\textbf{Model} & \textbf{ELO Rating} $\uparrow$ \\
\midrule
Hunyuan-V2A (w/ Caption)   & \textbf{1556} \\
MMAudio-Phys (w/ Caption)  & 1550 \\
ThinkSound (w/ Caption)    & 1509 \\
MMAudio (w/ Caption)       & 1447 \\
FoleyCrafter (w/ Caption)  & 1438 \\
\bottomrule

\end{tabular}
\end{table}

\subsection{Overall Analysis}
Our experiments collectively expose a fundamental bottleneck in modern V2A models: the video encoder. The results repeatedly show a central paradox that text captions improve semantic correctness but simultaneously degrade temporal alignment, pointing to a deep-seated architectural flaw. Current models treat text as the primary source of what to generate (the semantic category) while treating video as a secondary source of when to generate (the timing). The degradation in temporal alignment indicates a processing conflict: the model, in prioritizing text, tends to lose track of the precise visual cues for onset.

Conversely, the pronounced drop in semantic quality in the no-caption setting highlights this dependency. Most importantly, the consistently low physical confidence scores reveal that the video encoder, on its own, is not learning to extract physical properties from pixels. It cannot {\em see} the difference between metal and wood in a way that informs the audio synthesis. The model's limited understanding of physics is not an emergent property of visual simulation but is merely {\em parroted} from the text prompt when available.

We argue that the current V2A's focus must shift. The challenge is no longer just improving audio synthesis quality. The central problem is building video encoders that can internalize the rich physical and semantic information currently provided as a textual ``cheat''. Until this is solved, V2A models will remain naive script-readers, not true physical world models capable of reasoning from observation.

\section{Conclusion}
\label{sec:conclusion}
Our work introduces \dsetname, a new benchmark that reframes video-to-audio generation as a principled audit of a model's implicit understanding of the physical world. It moves beyond plausibility to audit causal responsiveness using controlled counterfactual interventions and single-video pattern tests. Our experiments reveal a consistent limitation: current models struggle to learn physical processes from pixels, with or without text conditioning. We find they are also deeply dependent on text for semantic understanding, and that this reliance creates a fundamental trade-off: semantic and physical correctness are typically gained at the cost of temporal synchronization. This work provides a tool to systematically measure this gap, reframing the central challenge for the field. The goal is no longer just about improving audio synthesis, but building visual representations that can internalize physical processes from pixels.

\mypar{Limitations and broader impacts.}
Our current benchmark is focused on single-factor interventions within indoor environments. It does not yet capture the complexity of compound physical interactions (e.g., simultaneous changes in force, material, and geometry) or the full spectrum of in-the-wild acoustic phenomena. We believe scaling this causal framework to more complex scenarios is a critical future direction. Furthermore, as models improve on these metrics, they will generate more causally convincing audio, which increases the potential for misuse in creating misleading media. Responsible development and detection strategies will remain essential.

\paragraph{Acknowledgements.}
We thank Yisi Liu, Yanshi Liu, Xinhui Li, and Mingxuan Cai for guidance and assistance with video recording. We also thank those nearby who put up with our many rounds of noisy recordings while we were collecting \dsetname.

{
    \small
    \bibliographystyle{ieeenat_fullname}
    \bibliography{main}
}

\clearpage
\appendix
\setcounter{page}{1}

\section{Project Webpage}
\label{app:proj_web}
We provide a \href{https://research.nvidia.com/labs/cosmos-lab/flatsounds/}{project webpage} that shows the physical correctness of current video-to-audio models through the following features:
\begin{itemize}[topsep=0pt, noitemsep, leftmargin=*]
    \item \textbf{Guided Physics Case Studies:} A set of curated examples from \dsetname that isolate specific physical factors, helping readers understand physically-grounded audio generation.
    \item \textbf{Counterfactual Pairs Showcase:} A demonstration of time-aligned video pairs where a single physical variable is manipulated. By warping videos to align impact timings, these examples allow users to directly attribute acoustic differences to the controlled factor, illustrating the benchmark's core methodology.
    \item \textbf{Interactive Video Explorer:} A filtering interface to browse and compare samples from all tested models, under both captioned and no-caption conditions.
    \item \textbf{Quantitative Visualizations:} Interactive plots, including a trade-off analysis scatter plot and a multi-metric radar chart, that provide a holistic view of model performance across physical, semantic, and temporal dimensions.
\end{itemize}

\section{Human Evaluation Details}
\label{sec:suppl_heval}
We set up human evaluation as a pairwise preference test to validate our benchmark metric, using a representative subset of 40 video clips from \dsetname. We chose pairwise comparisons because the alternative mean opinion score may not capture fine-grained differences between two models of roughly equal quality as each model's output is viewed in isolation. 

\begin{figure}
    \centering
    \includegraphics[trim=1.2px 0px 0px 5px,  clip,width=\columnwidth]{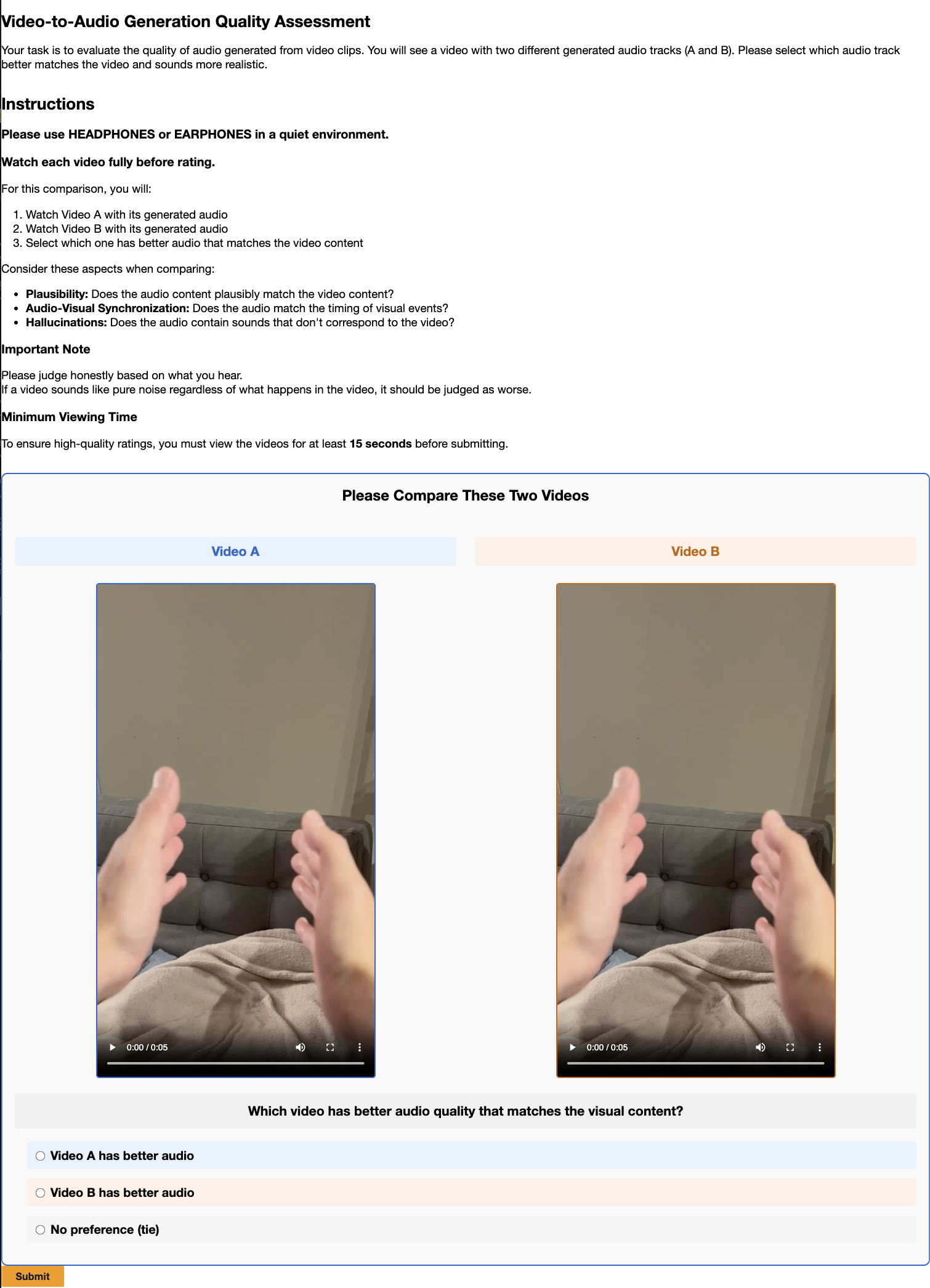}
    \caption{{\bf Example pairwise comparison page as presented to raters during the human eval study.} Each rater watches both videos and selects the one that best fits the specified criteria. We use a minimum viewing time to prevent raters from rating without watching. We also include a listening test by inserting a video paired with pure brown noise, which we ask the rater to always rate as less preferred, otherwise they will be rejected.}
    \label{fig:amt_screen}
\end{figure}

Due to the increased number of comparisons necessary for a pairwise setup, we limit our evaluation to the captioned versions of MMAudio-Phys, MMAudio, FoleyCrafter, Hunyuan-V2A, and ThinkSound. For each comparison, we first randomly select two models, one of the 40 videos to compare, and one of the 10 randomly generated audio tracks for each model, conditioned on the selected video.
The human rater is asked to select a preferred output based on plausibility, audio-video synchronization, and presence of auditory hallucinations. They are given the option to select the preferred output or ``no preference". We show an example comparison screen in Fig. \ref{fig:amt_screen}.

\begin{figure}
    \centering
    \includegraphics[width=\linewidth]{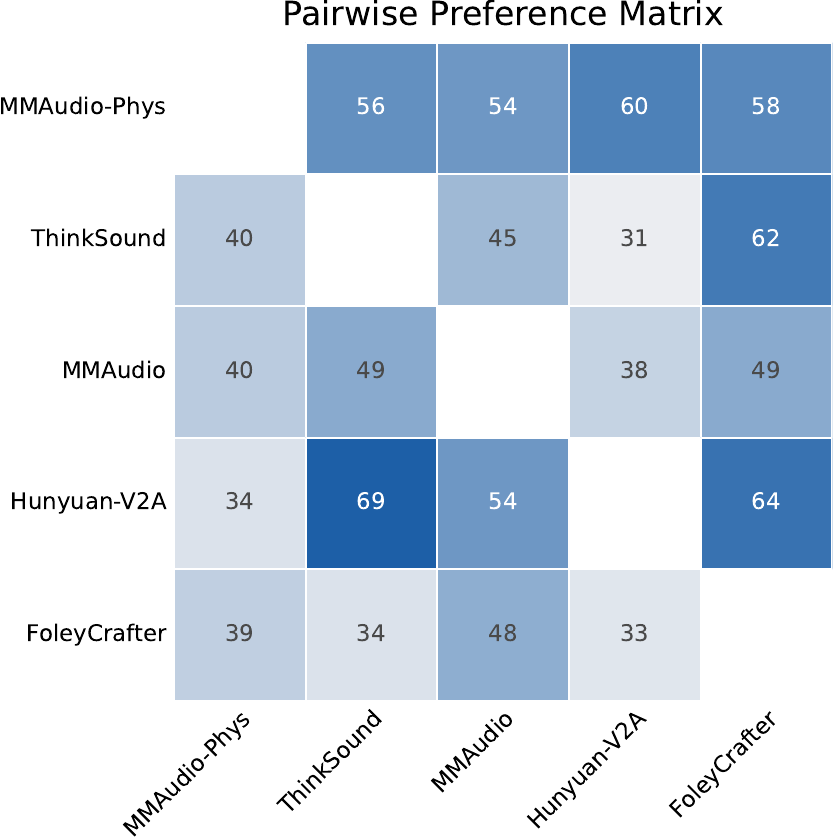}
    \caption{{\bf Win-rates for pairwise preference test.} Each cell $(i,j)$ represents how often the model for row $i$ is preferred over the model for column $j$. Overall, the pairwise preferences broadly support the ranking trends reported in the main paper's Human Evaluation discussion. We refer the reader there for the final ELO-based ordering.}
    \label{fig:heval_matrix}
\end{figure}

In total, we collected 578 valid pairwise comparisons from 382 workers (Fig. \ref{fig:heval_matrix}). The mean number of comparisons per worker was 1.5 with a median of 1. 31 workers were rejected for failing our noise test, where they must rate a video paired with brown noise as less preferred than the other video.

\mypar{ELO formulation.} We compute ELO ratings to produce a global ranking of models from pairwise comparison data as an efficient means of summarizing the pairwise results. The ELO system assigns each model \(i\) a rating \(R_i \in \mathbb{R}\), initialized to \(R_i = 1500\). After each pairwise comparison between model \(A\) and model \(B\), the ratings are updated based on the outcome.

Let \(R_A\) and \(R_B\) denote the current ratings of models \(A\) and \(B\). The expected score of model \(A\) is defined as:
\[
E_A = \frac{1}{1 + 10^{(R_B - R_A)/400}},
\]
and similarly for model \(B\):
\[
E_B = \frac{1}{1 + 10^{(R_A - R_B)/400}}.
\]
These values represent the probability that each model would win under the current ratings.

After observing the result, the ratings are updated using a fixed learning rate \(K = 32\):
\[
R_A' = R_A + K \cdot (S_A - E_A),
\]
\[
R_B' = R_B + K \cdot (S_B - E_B),
\]
where \(S_A, S_B\) are the outcome scores:
- \(S_A = 1, S_B = 0\) if \(A\) wins,
- \(S_A = 0, S_B = 1\) if \(B\) wins,
- \(S_A = S_B = 0.5\) in case of a tie.

This update is applied sequentially for all comparisons. The final ratings provide a latent performance score for each model, enabling a consistent global ranking even when not all models are directly compared.

\section{Hit Detection Details}
\label{sec:suppl_hit}
Here, we provide some more details on our setup for hit event detection from audio. Audio samples are first resampled to 44.1kHz, then we compute an STFT over the audio using a Hann window size of 1024 and hop size 256, and take the RMS along the time dimension of the STFT spectrogram to obtain a one-dimensional energy envelope.

\begin{figure}
    \centering
    \input{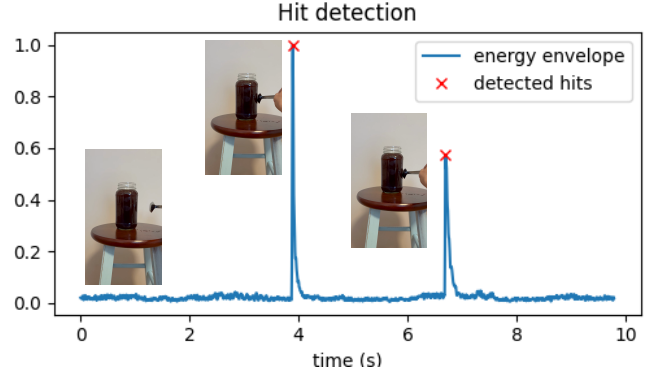}
    \vspace{-1em}
    \caption{Visualization of a clean example of the energy envelope peak detection used to annotate sound event timing.}
    \label{fig:hit_vis}
\end{figure}
\begin{figure}
    \centering
    \includegraphics[width=\linewidth]{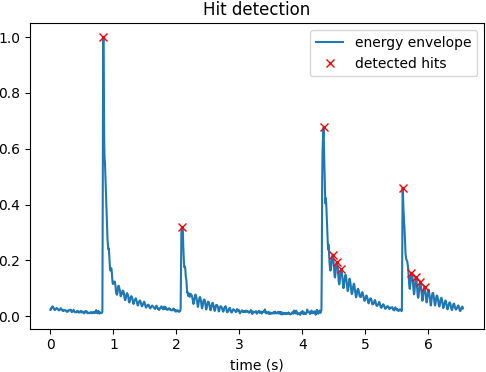}
    \caption{Example of a difficult case for automatically detecting sound events using our setup. The oscillations during the decay result in false positives.}
    \label{fig:hard_hits}
\end{figure}
However, peak detection performed this way is agnostic to anything other than the energy levels and is highly prone to false positives. Most of our cases look like the example in Fig. \ref{fig:hit_vis}, which shows a \emph{clink} sound of metal on glass against a relatively quiet backdrop and very distinct, clean peaks. However, we show an example in Fig. \ref{fig:hard_hits}, which corresponds to the sound of a metal spoon hitting a metal cup four times. The decay of the metal-on-metal sound oscillates in such a way as to generate multiple sub-peaks on its way down. We are able to automatically reject most of these by spacing out our ``hits'' roughly half a second apart, and rejecting anything within half a second of a detected peak.  However, it is difficult to tune this approach to be foolproof, as there are numerous valid sources of sounds other than those we annotated.
As such, we measure the coverage/recall of annotated hits in our evaluation.

\section{Dataset Distribution Details} 
\begin{figure*}
    \centering
    \includegraphics[width=\linewidth]{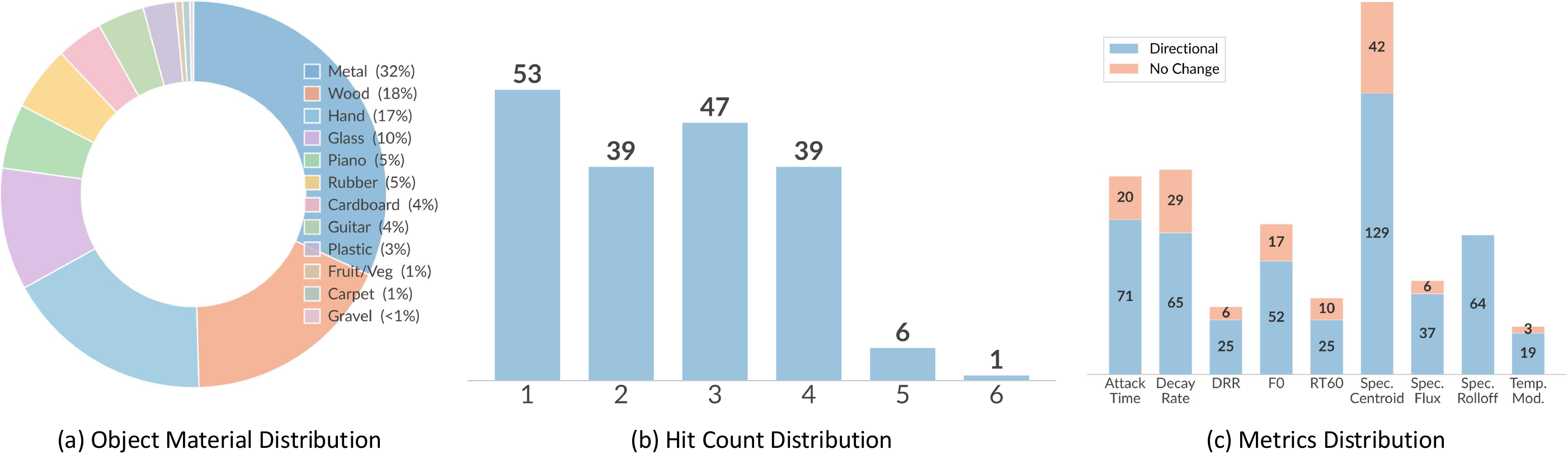}
    \caption{Distribution of (a) material types, (b) hit counts, and (c) metric-change annotations used to create counterfactual pairs. A pair may contain one or more metric change annotations, and the same object instance may appear in multiple videos, with each video featuring one or more interacting objects.}
    \label{fig:data_distribution}
\end{figure*}

\dsetname\ comprises 185 unique indoor clips. A pie chart for object material appearance distributions is shown in Fig. \ref{fig:data_distribution}, which shows the relative counts of each time an object of the specified material appears in a clip. Note that each object can appear in multiple clips, and most clips feature at least two interacting objects. Based on the pie chart, we can see that we heavily feature glass, metal, and wood. This is because we expect single-factor material changes between these materials to have relatively easy-to-detect differences with respect to our curated set of audio metrics. This is also reflected in the relatively large amount of counterfactual pairs with annotated expected changes in spectral-centroid as shown in Fig. \ref{fig:data_distribution}. Room metrics such as DRR and RT60, which measure reverberation levels of an acoustic environment, appear less frequently in our benchmark due to the relative difficulty in conveying a room's acoustic properties unambiguously within a video. We rely primarily on hallway/stairwell environments as settings where one would expect higher reverb, but we will likely look into other creative alternative approaches as we attempt to expand the size of our benchmark.

\section{Metrics Implementation Details}
\label{sec:metric_extraction_details}
\begin{figure}[t]
    \centering
    \includegraphics[width=\linewidth]{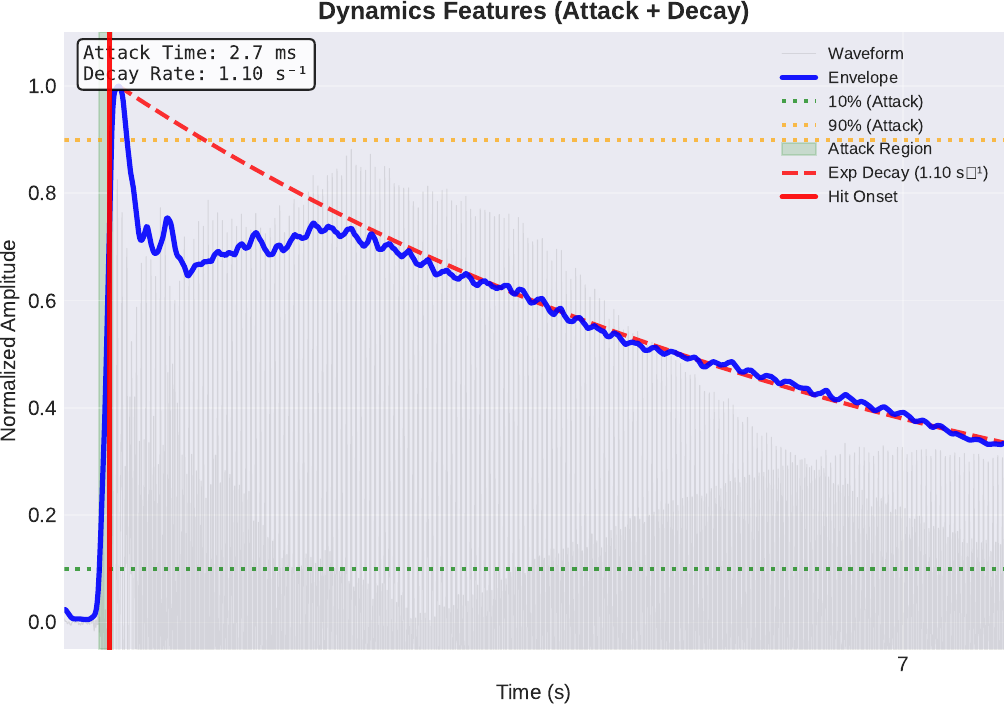}
    \caption{Plot showing the Attack time and Decay Rate of a detected hit.}
    \label{fig:temporal}
\end{figure}

\begin{figure}[t]
    \centering
    \includegraphics[width=\columnwidth]{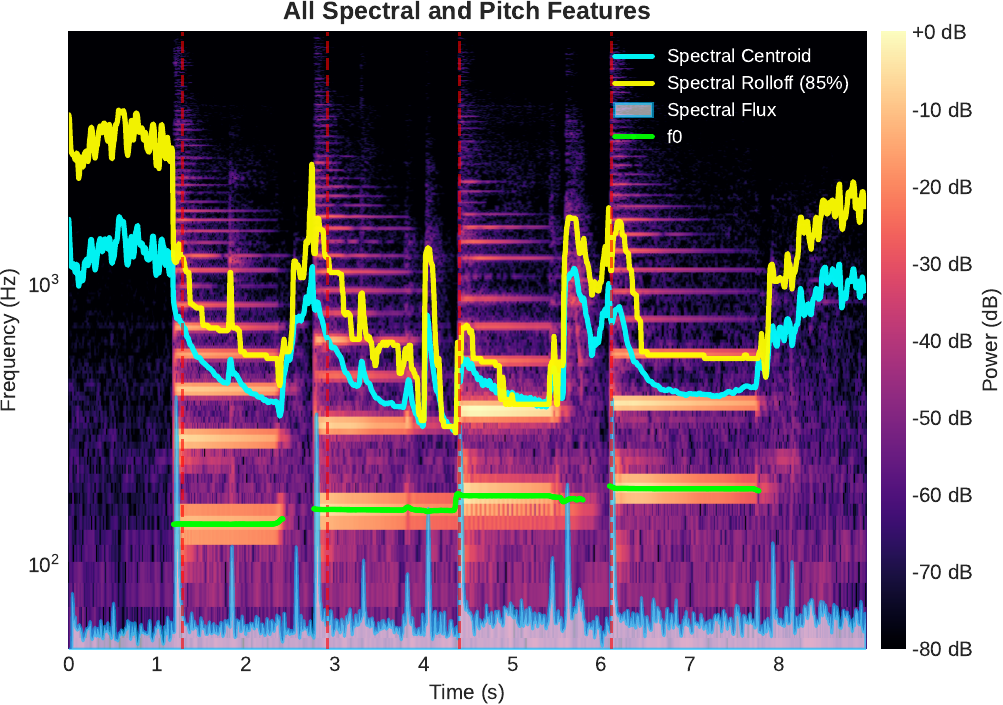}
    \caption{Plot showing all the spectral features, and the F0 contour overlaid on the audio spectrogram.}
    \label{fig:spectral}
\end{figure}

\begin{figure}[t]
    \centering
    \includegraphics[width=\columnwidth]{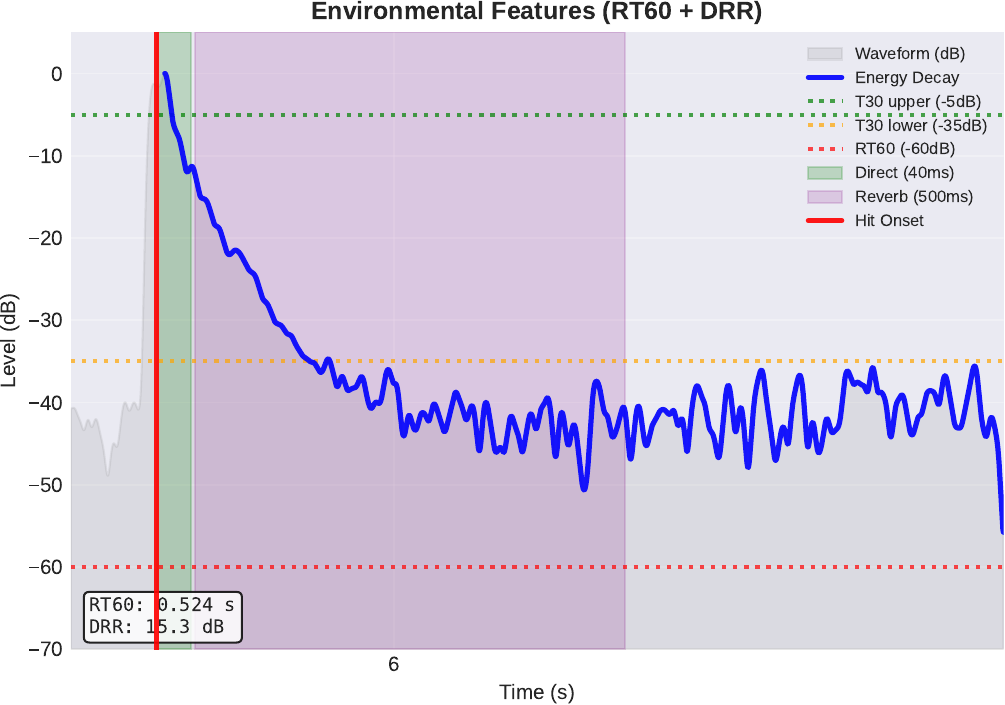}
    \caption{Plot showing RT60 and DRR measures of a detected hit.}
    \label{fig:env}
\end{figure}

Our benchmark evaluates physical correctness through nine acoustic metrics computed using robust signal processing techniques (Fig. \ref{fig:temporal}, \ref{fig:spectral}, and \ref{fig:env}). We employ a hybrid computation strategy: six metrics use \textit{per-hit averaging} (attack time, decay rate, fundamental frequency, spectral centroid, spectral rolloff, spectral flux), while temporal modulation is computed globally. RT60 and DRR remain room-acoustics metrics, but in multi-hit clips we estimate them in a hit-aware manner and aggregate them across hits. To ensure consistent spectral analysis across models with varying native sampling rates, all audio is resampled to 16kHz prior to analysis, with unified parameters (64ms analysis window and 8ms frame shift).

\subsection{Per-Hit Segmentation Strategy}

For multi-hit videos, the audio track is segmented at annotated hit times $\mathbf{t} = [t_1, t_2, \ldots, t_N]$. Each segment $s_i$ is defined to span from 50ms before the hit $t_i$ to the next hit $t_{i+1}$ (or the signal end). A default dynamic window adapted to the estimated RT60 is applied:
\begin{equation}
s_i = \text{audio}[\max(0, t_i - 50\text{ms}) : \min(t_{i+1} - 20\text{ms}, t_i + w_{\text{dur}})]
\end{equation}
Metrics are computed independently on each segment. For the per-hit metrics below, the resulting hit-level values are averaged across valid hits to obtain a video-level score.

\subsection{Attack Time}
We define the Attack Time as the duration required for the signal envelope to rise from 10\% to 90\% of its peak value. The computation process is detailed in Algorithm~\ref{alg:attack_time}.

\begin{algorithm}[ht]
\caption{Attack Time Calculation}
\label{alg:attack_time}
\begin{algorithmic}[1]
\State \textbf{Input:} Signal segment $y$, sampling rate $f_s$
\State $\text{env} \gets \text{GaussianSmooth}(|\text{Hilbert}(y)|, \sigma=3\text{ms})$
\State \textbf{Step 1: Onset Detection via Derivative Gating}
\State $\mu_{\text{noise}}, \sigma_{\text{noise}} \gets \text{Median}(\text{env}_{\text{pre}}), \text{MAD}(\text{env}_{\text{pre}})$
\State $\text{env}' \gets \nabla \text{env}$
\State $\sigma_{\text{der}} \gets \text{MAD}(\text{env}'_{\text{pre}})$
\State $i_{\text{onset}} \gets \text{First } i \text{ where } (\text{env}[i] > \mu_{\text{noise}} + 3\sigma_{\text{noise}}) \land (\text{env}'[i] > 3\sigma_{\text{der}})$
\State \textbf{Step 2: Peak Finding \& Monotonicization}
\State $i_{\text{peak}} \gets \arg\max_{i \in [0, 200\text{ms}]} \text{env}[i_{\text{onset}} + i]$
\State $\text{env}_{\text{mono}} \gets \text{CumulativeMax}(\text{env}[i_{\text{onset}} : i_{\text{peak}}])$
\State \textbf{Step 3: Rise Time Calculation}
\State $i_{10} \gets \min \{ i : \text{env}_{\text{mono}}[i] \geq 0.10 \times \text{peak} \}$
\State $i_{90} \gets \min \{ i : \text{env}_{\text{mono}}[i] \geq 0.90 \times \text{peak} \}$
\State \textbf{Return:} $1000 \cdot (i_{90} - i_{10}) / f_s$
\end{algorithmic}
\end{algorithm}

\subsection{Decay Rate}
We quantify the exponential energy dissipation of an impact event as $A(t) = A_0 e^{-\lambda t}$. The decay parameter $\lambda$ is estimated using a robust hierarchical fitting strategy described in Algorithm~\ref{alg:decay_rate}.

\begin{algorithm}[ht]
\caption{Decay Rate Estimation}
\label{alg:decay_rate}
\begin{algorithmic}[1]
\State \textbf{Input:} Signal envelope $\text{env}$
\State \textbf{Preprocessing:}
\State $\text{env}_{\text{norm}} \gets \text{CumulativeMin}(\text{env}[i_{\text{peak}}:] / \text{env}[i_{\text{peak}}])$
\State $\text{env}_{\text{dB}} \gets 20 \log_{10}(\text{env}_{\text{norm}})$
\State \textbf{Hierarchical Fitting (T30/T20/T10):}
\For{$\text{range} \in \{[-5, -35], [-10, -30], [-5, -25]\}$}
    \State $i_{\text{start}}, i_{\text{end}} \gets \text{FindCrossings}(\text{env}_{\text{dB}}, \text{range})$
    \If{$\text{Length}(i_{\text{start}}, i_{\text{end}}) \geq 6$}
        \State $m \gets \text{TheilSenRegression}(\text{env}_{\text{dB}}[i_{\text{start}}:i_{\text{end}}])$
        \If{$m < 0$ \textbf{and} $|m| > 10^{-6}$}
            \State $\lambda \gets -m / (20 / \ln 10)$
            \State \textbf{Return} $\text{Clip}(\lambda, 0.02, 50.0)$
        \EndIf
    \EndIf
\EndFor
\end{algorithmic}
\end{algorithm}

\subsection{Fundamental Frequency (F0)}
We employ a two-tier strategy to handle both harmonic and non-harmonic sounds robustly.
\begin{itemize}
    \item {\bf Parselmouth Autocorrelation:} We first extract a 300ms segment starting from the onset, discarding the initial 10ms transient. We compute the pitch using Praat-style autocorrelation \cite{jadoul2018introducing} within the range [27.5, 4186] Hz. We identify valid voiced frames $\mathcal{F}$ and compute the voiced ratio $r_{\text{voiced}}$. If $r_{\text{voiced}} \geq 0.1$ and at least three voiced frames are present, we compute F0 as the 10\% trimmed mean of $\mathcal{F}$. Additionally, we apply an octave error correction step: if $\text{F0} > 1200$ Hz, we check divisors $[2, 3, 4, 6, 8]$ and select the first candidate that falls within [80, 1500] Hz.
    \item {\bf Modal Peak Detection (Fallback):} If the voiced evidence is insufficient, we fall back to a spectral peak detection method. We extract a short 20-110ms window and compute the Welch Power Spectral Density (PSD) \cite{welch2003use}. We then detect peaks in the [80, 4000] Hz range that exceed an adaptive threshold $\tau = \text{median}(\text{PSD}) + 2.5 \times \text{MAD}(\text{PSD})$ and return the frequency of the lowest significant peak.
\end{itemize}

\subsection{Spectral Centroid \& Rolloff}
We compute both spectral metrics using a unified processing pipeline to capture the timbre of the impact's sustain phase \cite{lerch_introduction_2023}. We extract a fixed analysis window $y_{\text{timbre}}$ spanning from $t_{\text{onset}} + 60$ms to $t_{\text{onset}} + 180$ms to avoid the initial transient broadband noise. After removing the DC offset, we compute the Short-Time Fourier Transform (STFT) using an FFT size of $N_{\text{fft}}=1024$ and a hop size of 128 samples.

\mypar{Spectral centroid.}
We calculate the spectral centroid $\bar{f}[n]$ for each frame $n$ as the weighted mean of the component frequencies, representing the mass of the spectrum center:
\begin{equation}
\bar{f}[n] = \frac{\sum_{k=0}^{N_{\text{fft}}/2} f[k] \cdot |S[k, n]|}{\sum_{k=0}^{N_{\text{fft}}/2} |S[k, n]|}
\end{equation}
where $f[k]$ represents the center frequency of bin $k$, and $|S[k, n]|$ denotes the magnitude of the STFT at bin $k$ and frame $n$.

\mypar{Spectral rolloff.}
We compute the spectral rolloff $f_{85}[n]$ as the frequency threshold below which 85\% of the total spectral magnitude is contained:
\begin{equation}
\sum_{k=0}^{K_{85}} |S[k, n]| \geq 0.85 \sum_{k=0}^{N_{\text{fft}}/2} |S[k, n]|
\end{equation}
where $K_{85}$ is the bin index corresponding to $f_{85}[n]$.

Finally, we filter out invalid frames (e.g., silent frames resulting in division by zero) and aggregate the frame-wise values into a single scalar using a 10\% trimmed mean to ensure robustness against outliers.


\subsection{Spectral Flux}
We measure the strength of the attack transient by analyzing the rate of spectral change. An attack window of 180ms is extracted from the onset and RMS normalized. We then compute an onset-strength sequence,
\begin{equation}
\text{OS}[n] = \sum_{k=0}^{N_{\text{fft}}/2} \max(0, |S[k, n]| - |S[k, n-1]|)
\end{equation}
which serves as our spectral-flux proxy. We retain positive onset-strength values and apply Median Absolute Deviation (MAD) filtering to remove noisy frames \cite{leys2013detecting}. The final metric is the mean of the retained onset-strength values.

\subsection{Reverberation Time (RT60)}
We estimate the reverberation time using Schroeder integration \cite{schroeder1965new} combined with a hierarchical fitting approach.

\mypar{Energy decay curve.}
We first compute the Energy Decay Curve by integrating the squared signal energy backwards from the end of the signal to the peak amplitude $N$:
\begin{equation}
E[i] = \sum_{j=i}^{N} y[j]^2
\end{equation}
This curve is normalized and converted to dB scale. We also estimate the noise floor $\mu_{\text{noise}}$ from the last 10\% of the signal to determine a validity threshold.

\mypar{Hierarchical linear fitting.}
We attempt to fit a linear slope to the decay curve over three standard ranges: T30 ($-5$ to $-35$ dB), T20 ($-5$ to $-25$ dB), and T10 ($-5$ to $-15$ dB). For each range, segment length and duration are verified against minimum quality thresholds. Linear regression is performed to find the slope $m$, requiring $R^2 \geq 0.9$. RT60 is calculated as $-60/m$. If the T30 fit fails, the system progressively falls back to T20 and then T10.

\subsection{Direct-to-Reverberant Ratio (DRR)}
DRR calculation involves separating the signal into direct and reverberant components based on temporal windows. We define the direct sound window $w_{\text{direct}}$ as 40ms (for 16kHz audio) and adaptively set the reverberant window $w_{\text{reverb}}$ based on the estimated RT60.

We also apply a perceptual bandpass filter (125-4000 Hz) to both segments. The final DRR is computed as an energy ratio,
\begin{equation}
\mathrm{DRR} = 10 \log_{10}\left(\frac{\sum_n d[n]^2}{\sum_n r[n]^2}\right),
\end{equation}
where $d[n]$ and $r[n]$ denote the direct and reverberant segments, respectively. The result is clipped to the range $[-20, 40]$ dB.

\subsection{Temporal Modulation}
We combine three complementary measures to capture envelope variations: the Coefficient of Variation (CV), the Peak Factor (PF), and the Modulation Spectrum Energy ($E_{\text{mod}}$). We first extract the Hilbert envelope, apply anti-aliasing, and downsample it to approximately 200Hz.

\begin{itemize}
\item \textbf{CV:} Calculated as the standard deviation of the high-passed envelope divided by the mean of the downsampled envelope.
\item \textbf{Peak Factor:} Defined as the 99th percentile of the envelope divided by its RMS value.
\item \textbf{Modulation Spectrum:} We compute the FFT of the envelope ($P_{\text{mod}}$) and calculate the energy ratio in the 4-16 Hz band:
\begin{equation}
E_{\text{mod}} = \frac{\sum_{f=4}^{16} P_{\text{mod}}[f]}{\sum_{f>0} P_{\text{mod}}[f]}
\end{equation}
\end{itemize}

These three components are combined into a single weighted score, with the modulation-spectrum term heuristically upweighted to emphasize rhythmic structure:
\begin{equation}
\text{ModIndex} = 0.85 \times (0.4 \text{CV}_{\text{norm}} + 0.3 \text{PF}_{\text{norm}} + 0.6 E_{\text{mod}})
\end{equation}

\subsection{Robust Statistical Methods}
Throughout our pipeline, we employ robust statistics to ensure metric stability against noise and outliers. In particular, MAD-based thresholds and estimators are used extensively to suppress noisy frames and outliers \cite{leys2013detecting}. We also utilize Theil-Sen regression \cite{sen1968estimates} for slope estimation, which offers a breakdown point of 29\% compared to 0\% for ordinary least squares, making it highly effective for noisy decay curves.

\subsection{Onset Detection}
We implement a hybrid onset detection strategy to align generated audio with expected visual events. We primarily use an onset-strength-based detector with high temporal resolution ($\approx$3.3ms), and fall back to an envelope-based peak detector with adaptive thresholding when onset detection is unreliable. A greedy matching algorithm with adaptive temporal tolerance (100-250ms, depending on hit density) is then used to align detected audio onsets with ground-truth video timestamps.

\subsection{Temporal Alignment Metrics}
Once onsets are detected and aligned, we compute three metrics to quantify synchronization quality:
\begin{itemize}
    \item {\bf Hit Coverage:} This metric measures the recall of sound events. We consider a ground-truth event ``covered'' if at least one detected onset falls within an adaptive tolerance window around its annotated timestamp. Hit Coverage is simply the percentage of ground-truth events that are successfully covered.
    \item {\bf Timing Error:} For the subset of successfully covered events, we calculate the absolute temporal deviation between the ground-truth timestamp and the nearest detected onset. The Timing Error is reported as the mean deviation in milliseconds.
    \item {\bf Perfect Alignment:} To assess sequence-level consistency, we define Perfect Alignment as the percentage of generated clips where \emph{all} ground-truth events are successfully covered (i.e., 100\% Hit Coverage).
\end{itemize}

\subsection{Quality Weighting Framework}
We implement a quality weighting framework based on the premise that measuring fine-grained physical correctness is futile if the generated audio lacks basic temporal alignment or semantic accuracy. For example, analyzing the decay rate of a sound is meaningless if the sound onset is missed or if the generated class is incorrect. To address this, we down-weight the contribution of low-quality generations in our final confidence scores.

\mypar{Quality weight calculation.}
For each seed in a factual-counterfactual pair, we compute two scalar terms. The temporal term is defined as the minimum of the factual and counterfactual Hit Coverage scores for that seed. Similarly, the semantic term is defined as the minimum of the factual and counterfactual CLAP similarity scores for that seed. For single-video tests, the corresponding terms are taken from the single generated clip.

The final per-seed quality weight is the arithmetic mean of these two components ($w_i = 0.5 w_{\text{temporal},i} + 0.5 w_{\text{semantic},i}$). These per-seed weights are then incorporated into the voting procedure for the physical metrics while retaining all seeds in the denominator. This soft weighting approach allows us to preserve all samples in the evaluation while ensuring that the reported confidence reflects not just physical responsiveness, but also the fundamental quality of the generation.

\mypar{Dimension-aware handling.}
For seeds that fail to produce valid hits ($< 2$ detected hits), we set per-hit metrics (attack time, decay rate, F0, spectral centroid, rolloff, flux) to NaN and count them as failures in the denominator. However, room-acoustics metrics (RT60, DRR) and temporal modulation are still computed for these seeds, as they do not strictly require precise hit timing. All seeds, regardless of validity, count in the denominator when computing confidence scores.

\section{Physics-aware Caption Details}
To generate the physics-aware captions used in our study (e.g., for fine-tuning to get MMAudio-Phys), we employ a multi-stage pipeline for our audio-visual data. This process is designed to first capture modality-specific details from the audio and video streams independently, and then fuse them into a single physically grounded caption.

\subsection{Modality-Specific Caption Generation}

\paragraph{Audio captioning.}
First, we generate audio-only captions. This is accomplished by processing the soundtrack of each clip through the Qwen3-Omni model (Omni-Captioner) \cite{xu2025qwen3, ma2025omni} to produce a descriptive caption focused solely on the acoustic events. Note that this model does not accept text prompts and only takes audio as input.

\mypar{Video captioning.}
Concurrently, we generate visual-only captions from the silent video frames using Qwen3-VL \cite{qwen3-vl}. This model is guided by a comprehensive system prompt that enforces a visual-only analysis, with emphasis on physics, materials, and spatial reasoning. The specific prompts used for this stage are detailed below.

\begin{tcolorbox}[
    breakable,
    enhanced,
    colbacktitle=black,
    coltitle=white,
    fonttitle=\ttfamily\bfseries,
    fontupper=\ttfamily\small,
    title=System Prompt (Qwen3-VL),
    left=6pt, right=6pt, top=6pt, bottom=6pt
]
You are an advanced, visual-only AI specialized in generating descriptive captions for silent videos.
Work strictly from visual evidence (no audio; ignore OCR unless text is clearly legible) and be physics- and materials-aware.

\textbf{Objectives:}
\begin{itemize}
  \item Describe what is happening (people, objects, actions, motions, interactions).
  \item Capture spatial layout and physics-informed context (forces, constraints, material behavior).
  \item Produce immersive, human-readable captions for accessibility, indexing, and understanding.
\end{itemize}

\textbf{Guidelines:}

\textbf{Visual Content.}
\begin{itemize}
  \item Identify main subjects: people, animals, objects, vehicles, scenery.
  \item Describe actions and gestures precisely (e.g., ``a person raises a hammer and strikes'').
  \item Include salient visual details: clothing, facial expressions, colors, textures, clearly readable text/logos.
\end{itemize}

\textbf{Spatial / Physical Reasoning.}
\begin{itemize}
  \item Infer environment layout (indoor/outdoor, approximate space shape, obstacles, dominant surfaces).
  \item Use cues like shadows, reflections, lighting falloff, occlusion, deformation, motion trajectories.
  \item Indicate relative positions/distances/sizes when visually supportable.
  \item Note dynamics implying forces/constraints (accel/decel, recoil, friction/traction, elastic vs.\ inelastic contact, balance).
\end{itemize}

\textbf{Materials Awareness.}
\begin{itemize}
  \item When justified by visuals, identify likely materials (metal, wood, glass, plastic, fabric, stone, concrete, rubber, water/fluids).
  \item Base calls on visible cues: reflectance/specular highlights, translucency, grain/texture, deformation, wear/scratches, corrosion, residue/dust, joint types, fracture/splinter patterns, splash/flow/viscosity.
  \item If motion suggests material interaction (e.g., rubber skids on asphalt, glass shatters), briefly add that physical interpretation.
\end{itemize}

\textbf{Environmental Context.}
\begin{itemize}
  \item Name the setting when supported (workshop, office, street, kitchen, forest, stadium, etc.).
  \item Use indicators (walls, furniture, vegetation, weather, signage) to contextualize.
  \item Mention visible dynamics: dust, smoke, vibration, spray, lighting changes.
\end{itemize}

\textbf{Repetition \& Counting.}
\begin{itemize}
  \item If an action repeats and the repetitions are clearly distinguishable, include the count (e.g., ``knocks twice'', ``jumps three times'').
  \item If the exact count is uncertain (due to occlusion, motion blur, or ambiguity), do not give a number; use qualitative phrasing like ``repeatedly'', ``in quick succession'', or ``several times''.
\end{itemize}

\textbf{Inference Discipline (``two-cue rule'').}
\begin{itemize}
  \item You may include at most two cautious inferences in total that are not directly observed but are strongly suggested by $\geq 2$ visual cues each (optionally state the cues briefly).
  \item Example: ``likely steel (sharp specular reflections; no visible flex under load)''.
  \item If cues are insufficient or confidence is low, omit the inference.
  \item Never infer identities, ages, or private attributes; avoid lip reading or audio-based claims.
\end{itemize}

\textbf{Style \& Length.}
\begin{itemize}
  \item Clear, precise, natural language; concrete verbs over vague phrasing.
  \item Default to a longform caption: multi-sentence (and multi-paragraph if needed) with layered detail on actions, space, materials, and physical interactions.
  \item Target length: $\sim$150-300 words for typical clips; may extend for complex scenes. Avoid filler and speculation.
  \item Use approximate metrics only when visually justified.
\end{itemize}

\textbf{Output.}
\begin{itemize}
  \item Provide a longform caption (no time codes). If the user explicitly requests ``brief'', supply a shorter single-paragraph version.
\end{itemize}
\end{tcolorbox}

\begin{tcolorbox}[
    breakable,
    enhanced,
    colbacktitle=black,
    coltitle=white,
    fonttitle=\ttfamily\bfseries,
    fontupper=\ttfamily\small,
    title=User Prompt (Qwen3-VL),
    left=6pt, right=6pt, top=6pt, bottom=6pt
]
You will receive a silent video. Please generate a longform caption following the system rules, with emphasis on spatial layout, materials, physics-aware details, and counting clear repetitions.

Requirements for this video:
\begin{itemize}
  \item Visual-only analysis; do not assume any audio.
  \item Highlight spatial relationships, material properties, and physically plausible interactions.
  \item If an action repeats and repetitions are clearly visible, include the count; if the count is uncertain, describe it qualitatively without a number.
  \item Apply the two-cue rule: include up to two cautious inferences total, each supported by at least two visible cues; omit any inference you are not confident about.
  \item Default to a longform caption (approximately 150-300 words; longer if complexity warrants).
\end{itemize}

Output:
\begin{itemize}
  \item A longform caption with no time codes.
\end{itemize}
\end{tcolorbox}

\subsection{Audio-Visual Caption Fusion}
After generating the separate audio-only and video-only captions, we fuse them into a single, coherent, physics-aware caption. This fusion is performed by GPT-OSS \cite{agarwal2025gpt}. The model is given a highly constrained system prompt to act as a technical expert, synthesizing the two text inputs into an objective, audio-anchored, and physically grounded event log. The prompts for this fusion stage are provided below.

\begin{tcolorbox}[
    breakable,
    enhanced,
    colback=white,
    colframe=black,
    boxrule=0.4pt,
    arc=0pt,
    outer arc=0pt,
    left=6pt,
    right=6pt,
    top=6pt,
    bottom=6pt,
    colbacktitle=black,
    coltitle=white,
    fonttitle=\ttfamily\bfseries,
    fontupper=\ttfamily\small,
    title=System Prompt (GPT-OSS)
]
You are an expert AI for producing one objective, physically grounded technical caption by fusing two text-only inputs:
(1) an audio-only caption and (2) a video-only caption.

\medskip
\textbf{Primary Objective.}
Output only a concise, technical Final Caption that reads like an acoustic event log or a forensic analysis. Your task is to synthesize the most plausible physical event described by the two input texts. The Final Caption must:
\begin{itemize}
  \item infer object materials and environment from the textual evidence, embedding these facts directly and economically into the caption;
  \item precisely count and disambiguate distinct audio events and their physical causes as described in the captions;
  \item follow the constraint: no meta-reasoning, no analytical commentary, and no literary or poetic language; write only the objective physical description.
\end{itemize}

\medskip
\textbf{Length \& Structure (STRICT).}
\begin{itemize}
  \item Produce 80-100 words in English.
  \item Write in a single paragraph using clear, declarative sentences; avoid complex clauses and narrative storytelling.
  \item Focus on a clear chronological and causal sequence of events based on the fused information.
\end{itemize}

\medskip
\textbf{Information Fusion Protocol (STRICT).}
\begin{itemize}
  \item The two provided captions are your \textbf{only} source of information. Do not assume details not present in them.
  \item Synthesize a single, coherent description of the most physically plausible event consistent with both captions.
  \item Conflict resolution: if the captions conflict, prioritize the audio caption for acoustic details (what a sound is) and the video caption for spatial/object details (where an object is). Resolve contradictions through logical inference to create the most likely scenario. Do not mention the conflict itself or the source captions in the final output.
  \item Remove redundancy: if both captions mention the same event, merge the details into a single, efficient description.
\end{itemize}

\medskip
\textbf{Audio-Anchored Content Policy (STRICT).}
\begin{itemize}
  \item {Admission test for entities:} an entity from the captions is included in the final output only if it meets at least one of these criteria based on the input texts:
  \begin{enumerate}
    \item it emits sound; or
    \item it is in direct contact with a sounding source at the moment of sound; or
    \item it shapes the acoustics (reflects/absorbs/occludes/transmits) needed to explain a described sound; or
    \item it is the agent of a described sounding action.
  \end{enumerate}
  If none apply, exclude the entity from the final caption.
  \item {Silent-object cap:} at most two brief mentions of non-sounding context surfaces (e.g., ``concrete floor'', ``glass wall''). Use generic superclasses only (metal/wood/plastic/glass/stone/rubber/fabric), no colors, branding, micro-textures, or fine detail for silent items.
  \item {Sentence anchoring:} each sentence must contain a sound-linked action, count, acoustic cue, or causal link to a sounding event described in the captions.
  \item {Material inference scope:} infer or describe materials only for sounding objects or the immediate contact surfaces involved in those sounds, as suggested by the input texts.
\end{itemize}

\medskip
\textbf{Vocabulary \& Granularity.}
\begin{itemize}
  \item Controlled terms: metal (steel/aluminum), wood (solid/plywood), plastic (rigid/flexible), glass, ceramic, stone, concrete, rubber, fabric/textile, leather, paper/cardboard, water/ice.
  \item Composites allowed (e.g., ``wood tabletop with metal fasteners'').
\end{itemize}

\medskip
\textbf{Hedging Policy (STRICT).}
\begin{itemize}
  \item Do not use hedging words (``likely'', ``probably'', ``appears'', ``seems'', etc.). If uncertain about a detail due to conflicting or vague captions, fall back to a more generic description or omit the detail.
\end{itemize}

\medskip
\textbf{Style (STRICT, Technical Description).}
\begin{itemize}
  \item Third person, present tense.
  \item Tone must be clinical and direct, prioritizing physical accuracy over narrative flair. Avoid literary or poetic language.
\end{itemize}

\medskip
\textbf{Output Requirement.}
\begin{itemize}
  \item Output only the Final Caption paragraph, as specified above. No lists, no headings, no time-codes.
\end{itemize}
\end{tcolorbox}

\begin{tcolorbox}[
    breakable,
    enhanced,
    colbacktitle=black,
    coltitle=white,
    fonttitle=\ttfamily\bfseries,
    fontupper=\ttfamily\small,
    title=User Prompt (GPT-OSS),
    left=6pt, right=6pt, top=6pt, bottom=6pt
]
Here are the two pre-generated captions describing the same event.

Audio-only caption:
\{audio\_caption\}

Video-only caption:
\{video\_caption\}

Task:
Fuse the two captions to produce ONE Final Caption.

Requirements:
\begin{itemize}
    \item Language \& length: Write 80-100 words in English, in a single paragraph.
    \item Source of Truth: The provided captions are your only source of information. Synthesize the most plausible physical event they describe.
    \item Tone: The caption must be a technical, objective description of acoustic events and their physical causes. AVOID literary, poetic, or overly descriptive language.
    \item Content: Strictly follow the audio-anchored policy. Infer materials only for sounding objects or their immediate contact surfaces as suggested by the captions.
    \item Synthesis: Do not simply copy phrases from the input captions. Write a new, coherent description that logically integrates the information from both.
    \item No hedging or meta language. No lists, no headings, no time-codes.
\end{itemize}

Output:
Final Caption
\end{tcolorbox}

\begin{figure*}[t]
    \centering
    \includegraphics[width=\linewidth]{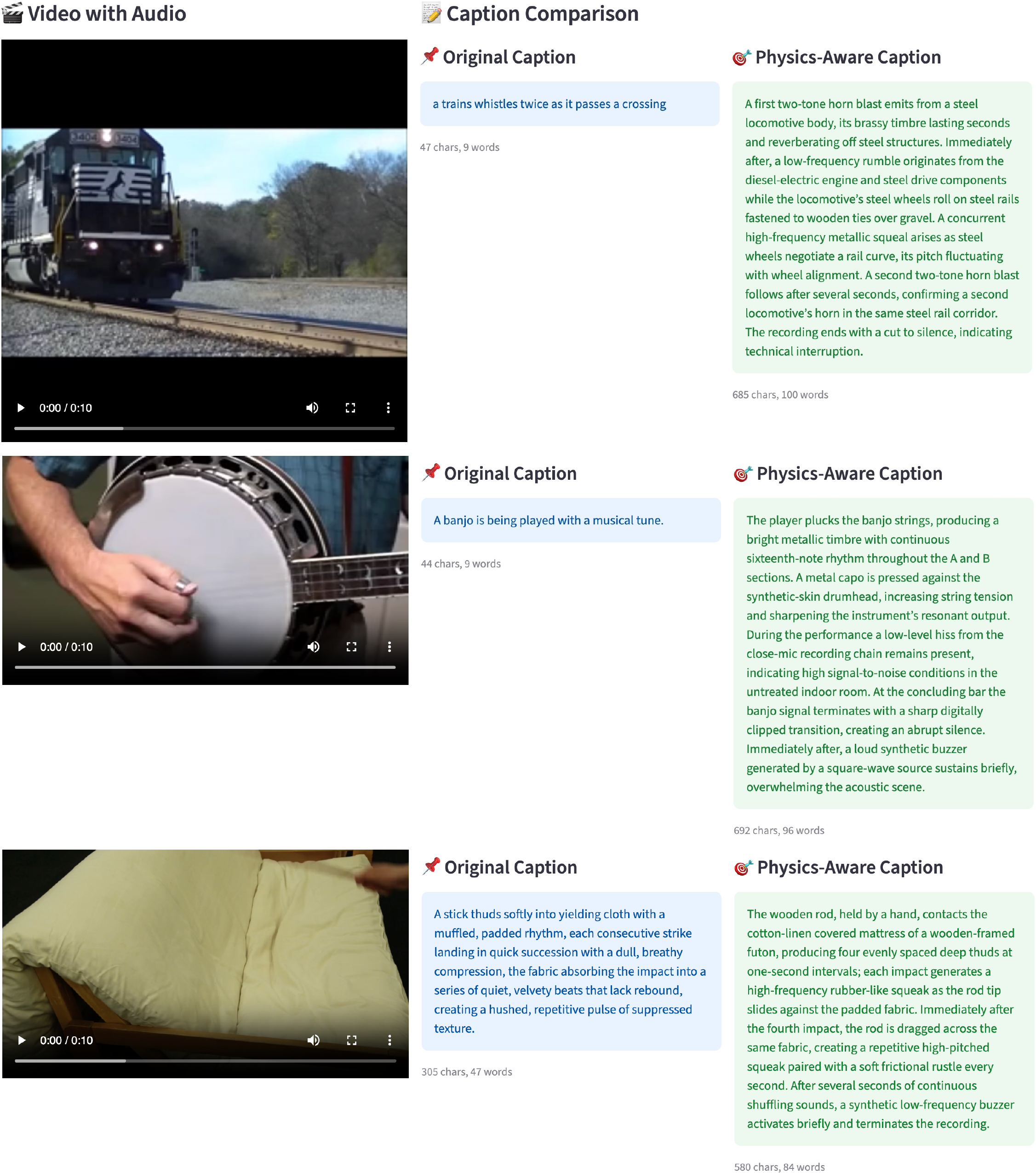}
    \caption{Qualitative comparison of original and physics-aware (our) captions for train (top), banjo (middle), and cloth (bottom) samples.}
    \label{fig:cap_comp}
\end{figure*}

\subsection{Caption Comparison}
To visually demonstrate the efficacy of our captioning pipeline, we compare original captions against our physics-aware captions. Fig.~\ref{fig:cap_comp} presents a qualitative comparison across three examples, sourced from AudioCaps \cite{kim2019audiocaps}, VGGSound \cite{chen2020vggsound}, and the Greatest Hits \cite{owens2016visually}.

Our generated captions explicitly ground acoustic events in their physical causes. For the train example (top), the caption captures dynamic acoustic shifts consistent with the Doppler effect and specifies material interactions like ``steel wheels'' on ``steel rails'', which is not covered in the original ``whistles twice''. Similarly, in the banjo clip (middle), the pipeline identifies the instrument's construction, explicitly citing the ``metal capo'' and ``synthetic-skin drumhead'' to explain the ``bright metallic timbre''. Finally, in the cloth example (bottom), our model details the contact dynamics, describing the ``high-frequency rubber-like squeak'' and ``frictional rustle'' of wood sliding against a mattress.

\section{MMAudio-Phys Training Details}
To test whether explicit physics-aware textual conditioning will improve semantic and physical correctness, we employ MMAudio \cite{cheng2025mmaudio} as the base architecture for our experiments. Our primary objective is not to propose a new model architecture, but to isolate the impact of enhanced caption quality on physical performance. MMAudio serves as an ideal test bed for this purpose due to its classic latent diffusion transformer design, accessible codebase, and state-of-the-art performance on standard benchmarks.

\mypar{Training data.}
As summarized in Table~\ref{tab:training_details}, our training dataset comprises about 3,360 hours of audio-visual and audio-text pairs sourced from 10 diverse datasets. To balance the data distribution, we oversample VGGSound samples by a factor of 3.

\begin{table}[t]
\scriptsize
\centering
\caption{Overview of datasets used in MMAudio-Phys.}
\label{tab:training_details}
\begin{tabular}{lcc}
\toprule
\textbf{Dataset} & \textbf{Type} & \textbf{Hours} \\
\midrule
WavCaps & Audio-Text & 1739.2 \\
FreeSound & Audio-Text & 625.6 \\
AudioSet & Audio-Text & 436.5 \\
VGGSound & Audio-Video & 405.8 \\
AudioCaps & Audio-Text & 117.8 \\
UrbanSound8K & Audio-Text & 18.9 \\
Greatest Hits & Audio-Video & 6.4 \\
ESC-50 & Audio-Text & 4.2 \\
MusicInstrument & Audio-Text & 4.1 \\
Sound of Water & Audio-Video & 2.1 \\
\midrule
\textbf{Total} & / & 3360.6 \\
\bottomrule
\end{tabular}
\end{table}

\mypar{Implementation details.}
We fine-tune the pre-trained MMAudio \texttt{large\_44k} checkpoint, which features a 21-layer transformer architecture (14 heads, 896 hidden dimension) operating on a latent sequence length of 345. The model processes 44.1kHz audio conditioned on both video features (CLIP: 64 frames, Synchformer: 192 frames) and text embeddings (1024-dim). Training is conducted with a total batch size of 256 for 700k iterations. We optimize using AdamW ($\beta_1=0.9, \beta_2=0.95$, weight decay $10^{-6}$) with a learning rate of $10^{-4}$. The schedule includes a 1k-step linear warmup, constant rate until 400k steps, and subsequent decays by $0.1$ at 500k and 600k steps. Classifier-free guidance is enabled by dropping input conditions with $p=0.1$, and exponential moving average (EMA, $\sigma_{\text{rel}} \in \{0.05, 0.1\}$) is applied throughout training. All training is performed using mixed-precision and the flow matching objective.

\section{Inference Setting}
For all tested models, we adhere to the default inference parameters specified in their official codebases. In the captioned setting, each test case is paired with one fixed descriptive caption that is shared across all methods to maintain a standardized inference interface. For ThinkSound, specifically, we use the provided chain-of-thought (CoT) prompts on the standard VGGSound benchmark. To ensure a fair evaluation across our \dsetname benchmark, we utilize a general CoT prompt for this model: \texttt{``Generate high-quality audio that matches the visual content''}. Notably, we observe that ThinkSound exhibits limited sensitivity to variations in the CoT prompt within the \dsetname benchmark. This behavior might stem from saturation in its training distribution, a phenomenon not observed in other evaluated models.

\end{document}